\documentclass{vgtc}                          

\ifpdf
  \pdfoutput=1\relax                   
  \usepackage{graphicx}                
  \DeclareGraphicsExtensions{.pdf,.png,.jpg,.jpeg} 
\else
  \usepackage{graphicx}                
  \DeclareGraphicsExtensions{.eps}     
\fi%
\graphicspath{{figures/}{pictures/}{images/}{./}} 
\usepackage{bm} 
\usepackage{subcaption}
\usepackage{amsfonts}
\usepackage{amsmath}
\usepackage{multirow}
\usepackage{inputenc}
\usepackage{microtype}                 
\PassOptionsToPackage{warn}{textcomp}  
\usepackage{textcomp}                  
\usepackage{mathptmx}                  
\usepackage{times}                     
\usepackage{cite}                      
\usepackage{tabu}                      
\usepackage{booktabs}                  
\usepackage{url}

\onlineid{6503}

\vgtccategory{Research}

\title{Learning to regulate 3D head shape by removing occluding hair from in-the-wild images}

\author{Sohan Anisetty\thanks{e-mail: sohananisetty630@gmail.com}\\ %
        \scriptsize IIT Dharwad %
\and Varsha Saravanabavan\thanks{e-mail: varsha005@ntu.edu.sg}\\ %
     \scriptsize NTU Singapore %
\and Cai Yiyu\thanks{e-mail: MYYCai@ntu.edu.sg}\\ %
    \scriptsize \centering NTU Singapore}

\abstract{Recent 3D face reconstruction methods reconstruct the entire head compared to earlier approaches which only model the face. Although these methods accurately reconstruct facial features, they do not explicitly regulate the upper part of the head. Extracting information about this part of the head is challenging due to varying degrees of occlusion by hair. We present a novel approach for modeling the upper head by removing occluding hair and reconstructing the skin, revealing information about the head shape. We introduce three objectives: 1) a dice consistency loss that enforces similarity between the overall head shape of the source and rendered image, 2) a scale consistency loss to ensure that head shape is accurately reproduced even if the upper part of the head is not visible, and 3) a 71 landmark detector trained using a moving average loss function to detect additional landmarks on the head. These objectives are used to train an encoder in an unsupervised manner to regress FLAME parameters from in-the-wild input images. Our unsupervised 3DMM model achieves state-of-the-art results on popular benchmarks and can be used to infer the head shape, facial features, and textures for direct use in animation or avatar creation.
} 

\CCScatlist{
  \CCScatTwelve{Human-centered computing}{Visu\-al\-iza\-tion}{}{3D morphable models}
}

\begin{document}


\firstsection{Introduction}

\maketitle
 

Blanz and Vetter~\cite{blanzvetter} introduced a general face representation and a principled approach to facial reconstruction. The key idea that any face can be defined as the linear combination of a set of template faces has given rise to 3D Morphable Models (3DMM). Traditionally, optimization-based analysis-by-synthesis methods have been used to infer the 3DMM parameters from a given image. However, this approach requires solving a complex and nonlinear optimization problem to model in-the-wild images. In recent years, morphable models have received increasing attention due to the advent of deep learning. Their nonlinear and deep representations offer a more robust and reliable performance on in-the-wild images (for a comprehensive overview of 3D morphable models see \cite{stateoftheart}). 


For practical use in 3D avatar creation and animation, we require reconstruction of the full head and not only the frontal face as computed by most of these existing models \cite{genova, microsoft, ganfit,3DMM_CNN}. Models like DECA\cite{DECA} and RingNet\cite{ringnet} that are based on the FLAME\cite{FLAME} geometric prior regress complete head shape. However, they train their models on frontal face features and do not account for the top part of the head due to occlusion by hair. Due to this, state-of-the-art models have high accuracy while evaluating with ground truth frontal face scans, but do not have a realistic head shape. As shown in Figure \ref{fig:DECA_compare}, DECA head reconstructions are enlarged compared to the input image. To address these issues we train an encoder that regresses 3DMM parameters from in-the-wild images. We regulate head shape using a dice consistency loss between masks of the input images and output renders. However, the skin mask of the input is not representative of the head shape due to a portion of the head being occluded by hair. To overcome this we develop deep learning-based models that can remove occluding hair and reconstruct the skin, making the person bald. We also quantitatively evaluate our model on the popular NoW benchmark\cite{NoWchallenge}, the CoMA\cite{coma} dataset, and LYHM\cite{LYHM} dataset.

\section{Related Work}



\subsection{3D reconstruction}

The existing monocular reconstruction algorithms can be broadly categorized into optimization-based \cite{totalmoving,face2face,dynamicavatar} and learning-based\cite{genova,microsoft,ganfit,ringnet,DECA}. Optimization-based methods aim to estimate 3DMM parameters by solving a non-linear optimization problem to minimize energy functions between the input and synthesized face image. Such methods are computationally expensive and output only a tightly cropped face mesh, unsuitable for animation. 


Learning-based methods utilize deep networks to regress 3DMM parameters. The use of non-linear representation for modeling surpasses classical models and can generalize well to unseen data. Inference speed is magnitudes faster than optimization methods, and training on a large corpus of data makes it robust to occlusions and varying real-world conditions. Learning-based models are usually trained in a supervised\cite{3DMM_CNN, prnet}, weakly supervised\cite{genova,microsoft}, or unsupervised\cite{DECA,ringnet,ganfit} fashion. Supervised learning models require 2D-3D paired data in the form of 3D scans or synthetic data. These methods do not generalize well to unseen data in the real world due to the low volume of training data that does not capture real-world complexity. Unsupervised approaches like \cite{ringnet,DECA,ganfit} regress 3DMM parameters without the need for explicit supervision. This form of learning eliminates the need for corresponding 3D data by instead minimizing loss functions between the input and rendered image. \cite{DECA} introduces shape consistency loss which enforces the fact that swapping only the shape parameters between images of the same subject should not affect the rendered image. Unlike our model, none of the existing methods, optimization or deep learning-based, take the upper head into consideration while designing their loss functions.

\subsection{Image editing}
Generative Adversarial Networks (GAN) \cite{GAN} learn the underlying patterns in the input data and generate new data that seem to be drawn from the real dataset with the help of an adversarial loss. Image manipulation using GANs gained prominence with the Pix2Pix framework \cite{pix2pix} which performed paired translation. However, this model requires paired training data, which is not readily available for hair to bald conversion. CycleGAN \cite{cyclegan} uses a novel cycle consistency loss to decipher the underlying style to perform the translation and requires an unpaired collection of photographs in each domain for training. 

Another popular method is the latent space exploration of GANs that manipulates images by altering their latent codes. First observed by \cite{interfacegan}, they succeed in manipulating gender, age, expression, and face pose by learning boundaries between the latent codes of these attributes. GAN inversion\cite{inversion_survey} is used to encode any image into the latent code of generative networks, extending latent space editing to in-the-wild images. Although latent space editing is successful on editing attributes such as age and hair color, they struggle with highly correlated attributes. 



Image inpainting is the task of filling the missing regions such that the filled pixels are semantically consistent with the rest of the image. Although Inpainting initially supported only regular polygon holes \cite{globallocal}, they later grew to support random masks with the help of edge generation \cite{edge_connect}, and partial convolutions\cite{partial_conv}. Recently some works have also used graph neural networks for image manipulation\cite{GNN_segment, Hypergraphs} due to their increased overall receptive field leading to better global consistency. The hypergraph-based inpainting approach\cite{Hypergraphs} is highly relevant to our purpose due to its ability to handle long-range contextual information.


\subsection{Quantitative evaluation}
The NoW benchmark\cite{NoWchallenge} focuses on extreme viewing angles, facial expressions, and partial occlusions. These methods evaluate error metrics only on a frontal crop of the head and not the entire head. However, to properly evaluate the accuracy of head shape reconstruction, we require $360^{\circ}$ scans of the subject's head. For this purpose we propose using the CoMA\cite{coma} dataset which provides FLAME registrations of 3D scans of 12 subjects in 12 challenging expressions, and the LYHM\cite{LYHM} dataset which provides FLAME registrations of 1000 subjects with a neutral expression.

\section{Approach}


The key idea is grounded in the observation that regardless of scale and crop, viewing angles, expression, and occlusion, the shape parameters must remain the same. Even if the upper part of the head is not visible in the input image, the model should accurately estimate the head shape. We enforce this idea by 
introducing a dice loss between skin masks of the input and rendered output of our model. However, the rendered output is bald, whereas the input has hair occluding the face leading to inconsistent skin masks and inaccurate dice coefficients. Thus, we remove hair from the input image to make a more fair and robust loss computation. In addition to the dice loss, we crop the input at multiple scale factors and shuffle the resulting shape parameters. This forces the encoder to infer complete head shape from facial features.






\begin{figure*}[]
\centering

\begin{subfigure}{.49\textwidth}
\includegraphics[ width = \textwidth , height = 4.5cm]{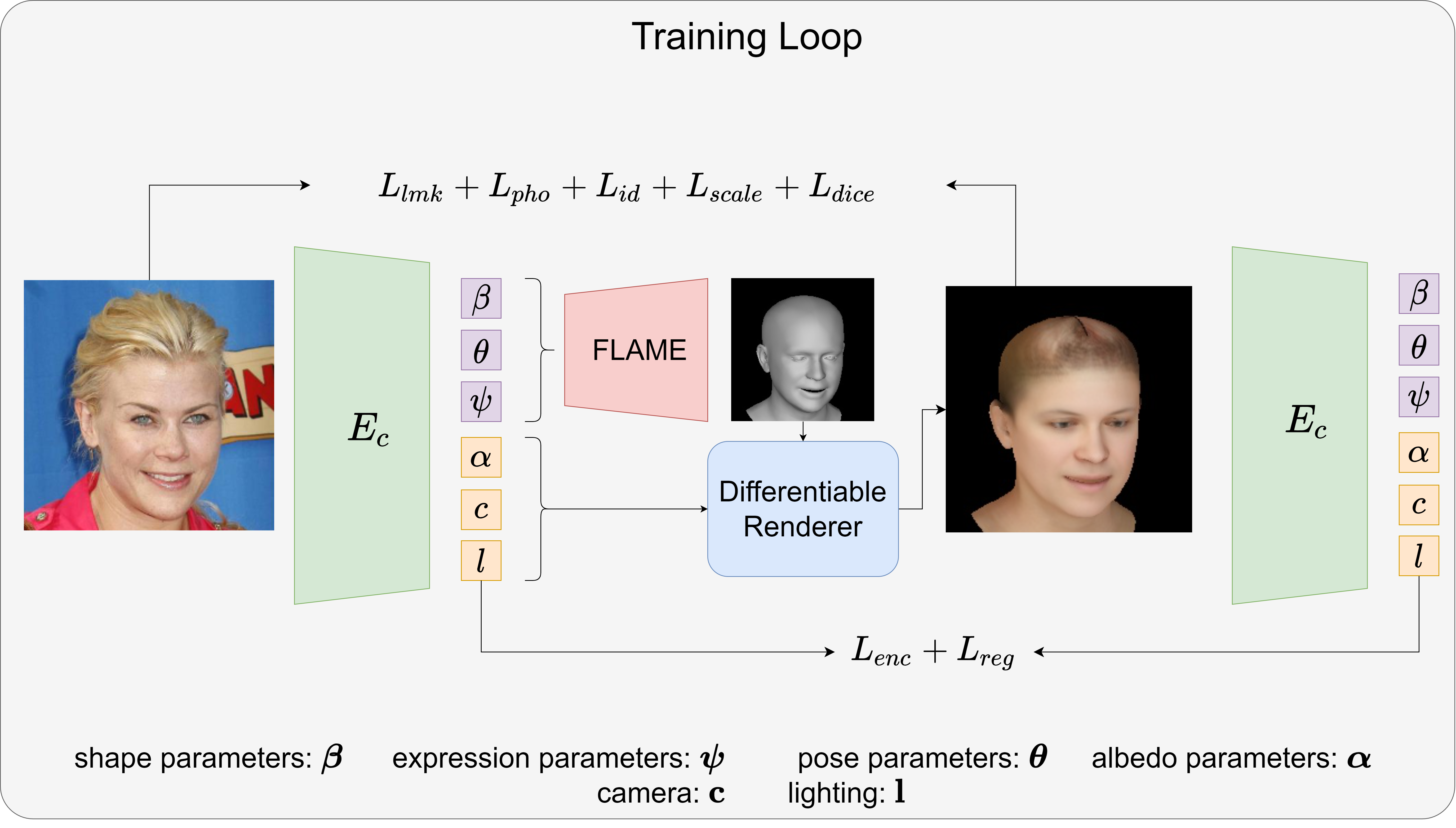}
\caption{}
\label{fig:training_loop}
\end{subfigure}%
\hfill
\begin{subfigure}{.49\textwidth}
\includegraphics[ width=\textwidth,height = 4.5cm]{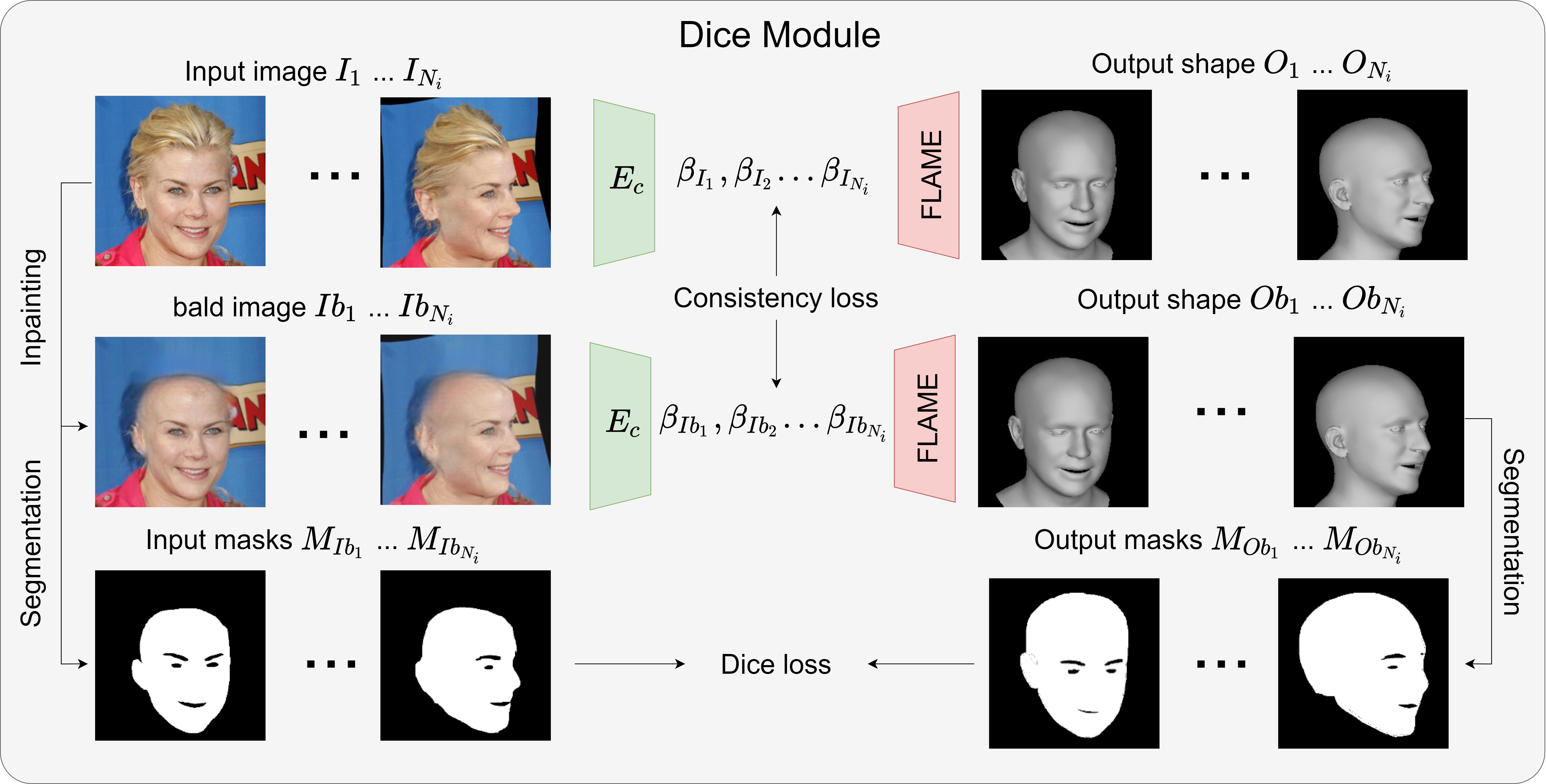}
\caption{}
\label{fig:dice_module}
\end{subfigure}%
\caption{The training pipeline (Fig. \ref{fig:training_loop}) showcases the flow of data; The input image is encoded to FLAME parameters using $E_c$. The FLAME decoder uses these encoded parameters to compute the geometric mesh that is textured and rendered using the differentiable renderer. The rendered image is passed through the encoder again to give parameters for calculating encoder loss. The Dice module (Fig. \ref{fig:dice_module}) outputs shape parameters $\beta_j$ corresponding to $N_i$ poses of input with hair ($I_j$) and without hair ($Ib_j$), where $j \in \{1,...N_i\}$. We generate skin masks of both the rendered ($M_{Ob_{j}}$) and input ($M_{Ib_{j}}$) images to compute the dice loss as given in Equation \ref{eq:dice}. We also minimize the difference between shape parameters of the input ($\beta_{I_{j}}$) and inpainted input ($\beta_{Ib_{j}}$). These shape parameters are also used by the scale module (Fig. \ref{fig:scale_module}) to calculate the scale consistency loss between parameters of $N_s$ images of the subject cropped in various ways using Equation \ref{eq:scale}. }
\label{fig:TrainingLoop}
\end{figure*}


\subsection{Hair removal}


We opt for a deep learning approach using generative adversarial models for removing hair on top of the head. We first experiment with popular image translation methods, then with latent space editing, and finally with inpainting methods as described in this section. The results of these methods are consolidated and compared in Section \ref{Hair_removal_result}. 

\subsubsection{Image translation}

\textbf{Pix2pix: }Pix2Pix\cite{pix2pix} requires paired examples to learn the relation between input and output. Due to non-availability of datasets with hair-bald paired examples, we create an artificial dataset in a commercial 3D modeling software\cite{blender} and train the Pix2Pix network on this synthetic dataset. This network
architecture comprises of a PatchGAN discriminator with a U-Net generator. We also augment training with an additional identity loss to preserve facial characteristics during translation.


\noindent \textbf{CycleGAN: }The CycleGAN model is trained on an unpaired collection of photographs from each domain(subjects with hair and without) taken from the CelebA dataset\cite{celebA}. This network consists of a PatchGAN discriminator and a generator with encoder-decoder architecture. We also introduce an identity loss to preserve facial characteristics during translation.

\subsubsection{Latent space editing}
For each unique d-dimensional latent code $\textit{z} \in \mathbb{R}^d$, a GAN generator \textit{G} generates a unique image $\textit{G(z)}$. \cite{interfacegan} observed that this \textit{z} vector encodes information about facial characteristics, and modifying the \textit{z} vector would result in the modification of those characteristics in the image. Given a separating hyperplane with unit normal vector $\textit{n} \in \mathbb{R}^d$ between latent codes corresponding to specific characteristics, we estimate the perturbation magnitude $\epsilon$ such that $\textbf{z}_{new} = \textbf{z} + \epsilon\textit{n}$, where  $\textbf{z}_{new}$ corresponds to latent code with desired attribute. We estimate $\epsilon$ by trying to increase the predicted score $\alpha = f(g(\textbf{z}_{new}))$ of desired attribute, where \textit{f} is a Resnet\cite{resnet} based multi-class classifier which predicts probabilities of different attributes present in the image. In our case, we try to maximize the attribute of baldness. We use \cite{indomain} to obtain latent codes of images of people with hair and without. This set of latent codes was used to train a linear SVM  resulting in the desired separating hyperplane. We use \cite{indomain} to invert the input image into the latent space of StyleGAN\cite{stylegan}. To further improve the quality of the inversion we introduce the M-SSIM loss\cite{msssim} which additionally gives a quantifiable metric to measure inversion accuracy. 





\begin{figure}[]
\centering
\includegraphics[height = 3.6cm]{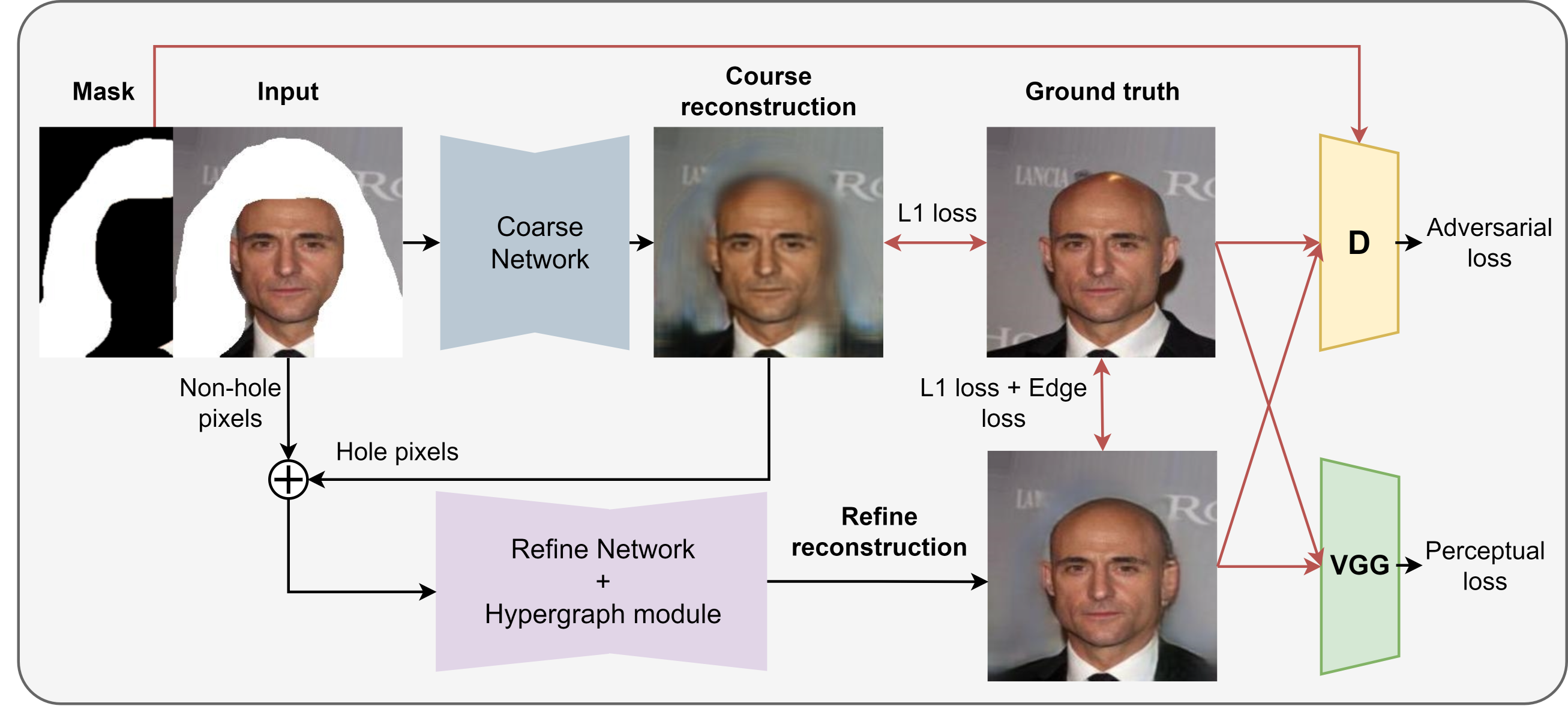}
 \caption{ The inpainting pipeline takes a masked image as input and reconstructs skin using a course network and a refine network consisting of  hypergraph modules. The black arrows signify data flow while the red arrows show loss computations during training.
}
\label{fig:inpainting_pipeline} 
\end{figure}

\subsubsection{GAN inpainting}

We utilize the hypergraph network architecture\cite{Hypergraphs} for inpainting due to the fact that complex relationship among spatial features of an image can be represented effectively by hyperedges that can connect more than two nodes i.e. spatial features, which a traditional CNN struggles to do due to its limited receptive field. Our task is to remove hair using this network, therefore, we denote hair as the missing region or hole which the inpainting network fills with valid pixels(reconstructed skin and background). Given a source image \textit{I} and a binary mask $\textit{M}$ of the missing region(1 for holes), the inpainting network fills in the missing pixels and gives as output a coarse reconstruction $I_{coarse}$ further refined using hypergraph layers to give $I_{refine}$ using a course and refine network respectively. $I_{refine}$ is the final output with hair removed and skin reconstructed to be used for training the 3DMM model. This inpainting network is trained on a dataset of bald people overlayed with random hair masks generated from subjects with hair in the CelebA dataset. The training pipeline is shown in Figure \ref{fig:inpainting_pipeline}. Formally, we minimize 




\begin{equation}
L_{total} = L_{hole} + L_{valid} + L_{adv} + L_{ppl} + L_{edge}.
\end{equation}
Where,

$L_{hole}$ and $L_{valid}$ are photometric losses to enforce consistency in the hole region and non-hole region respectively, $L_{adv}$ is the adversarial loss between the generator(refine network) and discriminator of the GAN model, $L_{ppl}$ is the perceptual similarity metric\cite{lpips}, and $L_{edge}$ is the edge consistency loss\cite{edge_connect}. These loss functions are defined as,

\begin{multline*}
L_{hole} = || M \odot (I_{refine} - I_{gt} ) ||_1 + || M \odot (I_{coarse} - I_{gt} ) ||_1 \\
L_{valid} = || (1-M) \odot (I_{refine} - I_{gt} ) ||_1 + || (1-M) \odot (I_{coarse} - I_{gt} )||_1
\end{multline*}

$L_{hole}$ is the loss on the hole pixels values, and $L_{valid}$ is the loss on the non-pixels values. The adversarial loss $L_{adv}$ can be formulated as a min-max problem between the inpainting generator network \textit{G} which predicts $I_{refine}$  and the discriminator \textit{D}.

\begin{equation}
L_{adv} = \mathop{max}_{D} \mathop{min}_{G} \mathbb{E}[log(D(I_{gt} , M))] + \mathbb{E}[log(1 - D(G(I_{in}) , M))]
\end{equation}
\cite{lpips} demonstrated the effectiveness of deep features as a perceptual similarity metric compared to traditional ones like PSNR\cite{psnr} which fail to account for many nuances of human perception. The perceptual loss is used to encourage a reconstruction to appear more natural to the human eyes. Let $\phi_l$ denote features of the $l^{th}$ activation layer of a pre-trained network, the perceptual loss is,
\begin{equation}
L_{ppl} = \sum_l || \phi_l(G(I_{in})) - \phi_l(I_{gt}) ||
\end{equation}
We evaluate this loss for the final prediction $I_{refine}$ and $I_{comp}$ ( $I_{refine}$ with valid pixel set to corresponding pixels of $I_{gt}$\cite{partial_conv} ). We also use a edge-preserving loss\cite{edge_connect} using the sobel filter \textit{E},
\begin{equation}
L_{edge} = || E(I_{refine}) - E(I_{gt}) ||
\end{equation}

\subsection{Landmark detection}
Following FAN\cite{FAN} which demonstrated the impressive power of stacked hourglass networks\cite{stackedhourglass} for landmark detection, we train a two-stack hourglass network to detect 71 landmarks. Due to the non-availability of annotated data, we manually annotate 2,000 images with three landmarks on top of the head in addition to 68 landmarks detected by current models. Due to the lower volume of training data, we train the model incrementally using an exponentially moving average loss function with greater weight given to the latest landmark. We use this detector to train our 3DMM model. We then use this trained 3DMM model to infer additional landmarks on the 300WLP dataset and use this to further refine the landmark detector.

\subsection{Monocular reconstruction}

We learn to regress FLAME parameters from a single image using an analysis-by-synthesis approach. Given a 2D image \textit{I} as input we encode the image into FLAME parameters and use it to synthesize a rendered image $I_r$ using
the FLAME decoder and a differentiable renderer, and minimize the error between this image and the input image as shown in Figure \ref{fig:training_loop}. The FLAME parameters consists of 100 shape parameters $\bm{\beta}$, 50 expression parameters $\bm{\psi}$, 6 pose parameters $\bm{\theta}$, 50 albedo parameters $\bm{\alpha}$, 3 camera parameters \textbf{\textit{c}}, and 27 lighting \textbf{\textit{l}}  coefficients. The goal of the reconstruction network is to minimize,
\begin{equation}
L_{full} = L_{lmk} 
+L_{pho} + L_{id} + L_{scale} +L_{dice} + L_{enc} + L_{reg},
\end{equation}
with landmark loss $L_{lmk}$, 
photometric loss $L_{pho}$, identity loss $L_{id}$, scale consistency loss $L_{scale}$, dice consistency loss $L_{dice}$, encoder loss $L_{enc}$, and regularization $L_{reg}$.\\

\noindent \textbf{Landmark loss: } The landmark loss measures the Manhattan distance between ground truth 2D face landmarks $\textbf{l}_i$ and corresponding 3D landmarks $\textbf{l}^{\prime}_i$ on the mesh surface projected on the image plane with weights $\omega_i$.
\begin{equation}
L_{lmk} = \sum_{i=1}^{71} \omega_i ||\textbf{l}_i - \textbf{l}^{\prime}_i ||_1
\end{equation}
\\



\noindent \textbf{Photometric loss: }The photometric loss computes the discrepancy between the input image \textit{I} and the rendered image $\textit{I}_r$ in regions specified by a binary skin mask $\textit{M}_s$ as,

\begin{equation}
L_{pho} = || \textit{M}_s \odot (\textit{I} - \textit{I}_r)||_1,
\end{equation}
where $\odot$ denotes the Hadamard product. This skin mask is computed using a segmentation network\cite{faceparsing} built upon a Bisenet\cite{bisenet} architecture and trained on the CelebAMask-HQ\cite{celebAMaskHQ} dataset. Using the skin mask prevents the influence of occlusions on the loss improving the quality of the reconstructed texture.\\

\noindent \textbf{Identity loss: }Regardless of expression, pose or illumination, embeddingsfrom the face recognition network\cite{vggface2} \textbf{\textit{f}} map close to each other for the same identity. We quantify the similarity between the input \textit{I} and rendered image $\textit{I}_r$ as the cosine score between their embeddings $\textit{f}(\textit{I})$ and $\textit{f}(\textit{I}_{r})$. We define the loss as
\begin{equation}
L_{id} = 1 - \frac{f(I)f(I_r)}{||f(I)||_2 \cdot ||f(I_r)||_2}.
\end{equation}

\begin{figure}[]
\centering

\begin{subfigure}{.65\columnwidth}
\includegraphics[ width = \columnwidth , height = 3cm]{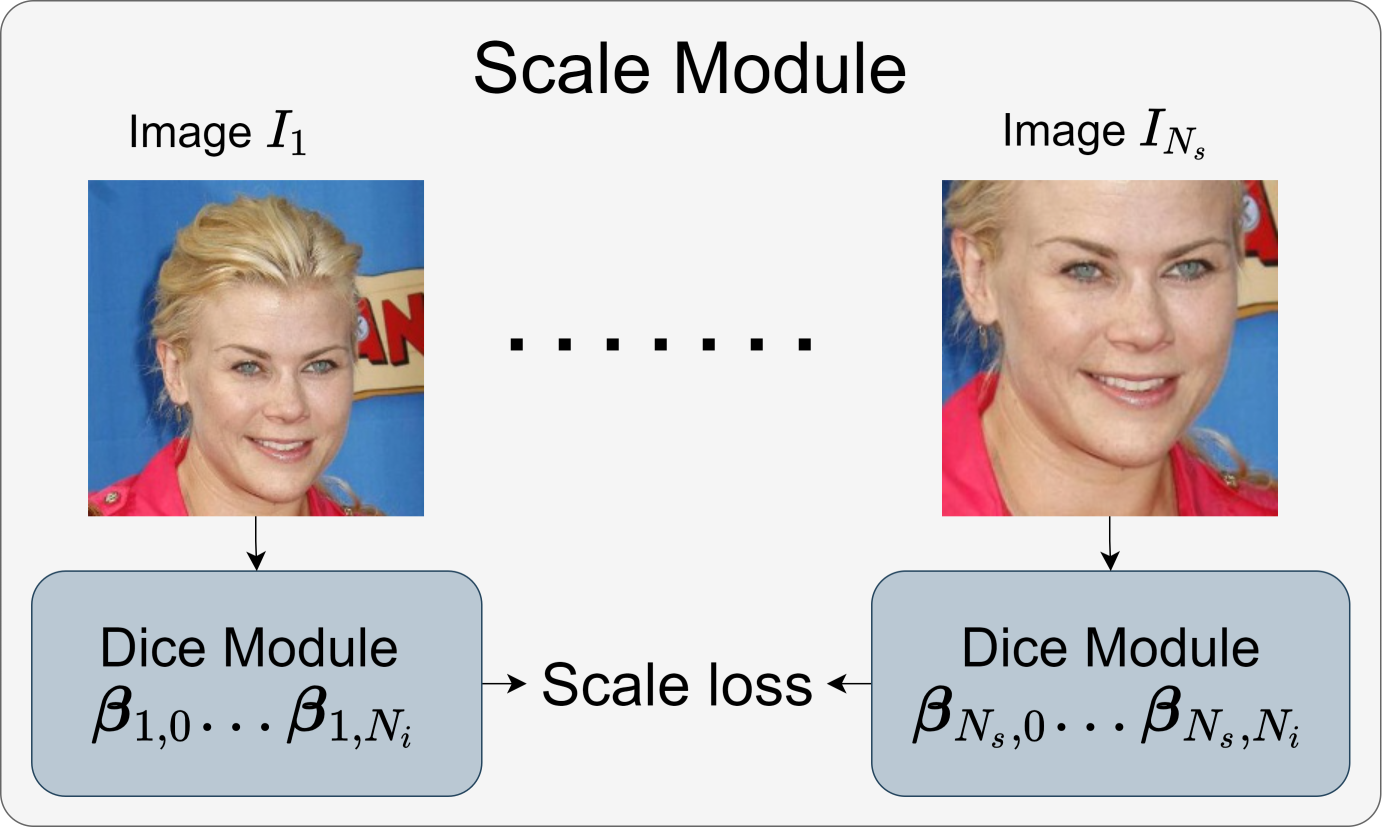}
\caption{}
\label{fig:scale_module}
\end{subfigure}%
\hfill
\begin{subfigure}{.18\columnwidth}
\includegraphics[ width=\columnwidth,height = 3cm]{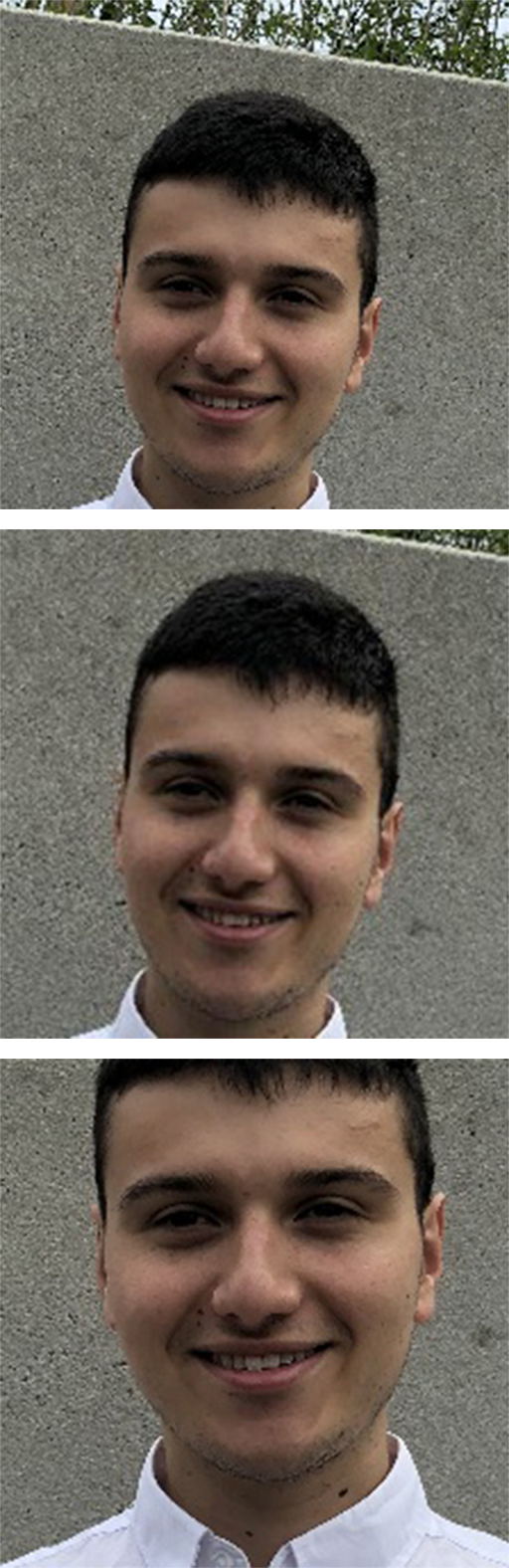}
\caption{}
\label{fig:cropLevels}
\end{subfigure}%

\caption{The scale module (\ref{fig:scale_module}) enforces similarity between shape parameters $\beta$ of $N_i$ images of a person in varying poses across $N_s$ scales. Fig. \ref{fig:cropLevels} shows an example of $N_s = 3$ levels of scale, from top to bottom level 3, level 2 and level 1. }
\label{fig:scale_loss}
\end{figure}

\noindent \textbf{Scale consistency loss: }Similar to DECA's shape consistency and RingNet's ring loss, we use multiple images of the same person to isolate shape parameters from expression and pose. In the scale consistency loss we crop the input image in $N_s$ different way 
ranging from a part of the head not being visible, to the entire head being visible as shown in Figure \ref{fig:cropLevels}. These crops are used in the scale consistency loss defined below in tandem with multiple views of the same person as shown in Figure \ref{fig:scale_module}.  Given $N_i$ images of the same person in different poses, we obtain $N_s$ crops of each image resulting in $N_i \cdot N_s$ shape parameters for a single subject. A scale-invariant encoder should output the same shape parameters $\beta$ irrespective of image conditions. Therefore, we shuffle the shape parameters of the same identity, while keeping other parameters constant and enforce the fact that this should not affect the rendered images. For image $I_{ik}$:

\begin{equation}
\label{eq:scale}
L_{scale} = L_(\textit{I}_{ik}, \textbf{R}( \textbf{F}(\beta_{{jl}}, \psi_{{ik}}, \theta_{{ik}}) , \textbf{A}(\alpha_{{ik}}, l_{{ik}}) , \textbf{c}_{{ik}} )     ),
\end{equation}
where $i,j \in N_i$ and $k,l \in N_s $, $\textbf{R}$ represents the renderer, $\textbf{F}$ the FLAME geometric model and $\textbf{A}$ the appearance model, and \textit{L} is a combination of landmark, photometric, identity and encoder loss. This allows the model to decipher the upper head structure based on the facial features.\\

\noindent \textbf{Dice consistency loss: }
By training the model simultaneously on the input with hair and without it, the model learns to decipher the upper head shape during inference. This is achieved using a combination of dice loss and consistency loss.
 
 
\textbf{Dice loss: }Given an input image \textit{I}, we remove hair using the inpainting network resulting in an image with the hair removed $\textit{I}_{b}$ and pass it through the segmentation network to obtain skin mask $M_{Ib}$. The dice loss between skin mask  $M_{Ib}$ of input and skin mask $M_O$ of the rendered output is defined as 

\begin{equation}
dice\ loss = 1 -  \frac{2 M_{Ib} M_O + \epsilon}{ M_{Ib} +  M_O + \epsilon},
\label{eq:dice}
\end{equation}
where we use $\epsilon$ to ensure that the function is not undefined when $M_{Ib} = M_O = 0$. 

Since during inference we do not perform inpainting, we design a parallel evaluation scheme as shown in Figure \ref{fig:dice_module} where we minimise the difference between shape parameters of the input and it's bald counterpart. The consistency loss is simultaneously calculated between \textit{I} and $\textit{I}_{b}$.

\textbf{Consistency loss: }We pass \textit{I} and $\textit{I}_{b}$ through the encoder $\textit{E}_c$ giving shape parameters $\beta_I$ and $\beta_{Ib}$. We enforce the fact that these shape parameters should be equal by shuffling $\beta_I$ and $\beta_{Ib}$ and minimizing the loss between rendered image and input image similar to Equation \ref{eq:scale}.\\


\noindent \textbf{Encoder loss: } The previous metrics enforced similarity between the rendered and input image. We also minimize the difference between the encoded FLAME parameters of the rendered image $\textit{p}_{out}$ and the input image $\textit{p}_{in}$. 

\begin{equation}
L_{enc} = \sum_{p \in \{\beta, \psi,\theta,\alpha,c,l\} } ||p_{in} - p_{out}||^2_2,
\end{equation}

\noindent \textbf{Regularization: } We define the regularization loss on shape $\beta$, expression $\psi$, and albedo $\alpha$ parameters as
\begin{equation}
L_{reg} = ||\beta||^2_2 + ||\psi||^2_2 + ||\alpha||^2_2
\end{equation}

\section{Evaluation}

\subsection{Qualitative evaluation}

\begin{figure}[]
\centering
\includegraphics[width = 4.5cm]{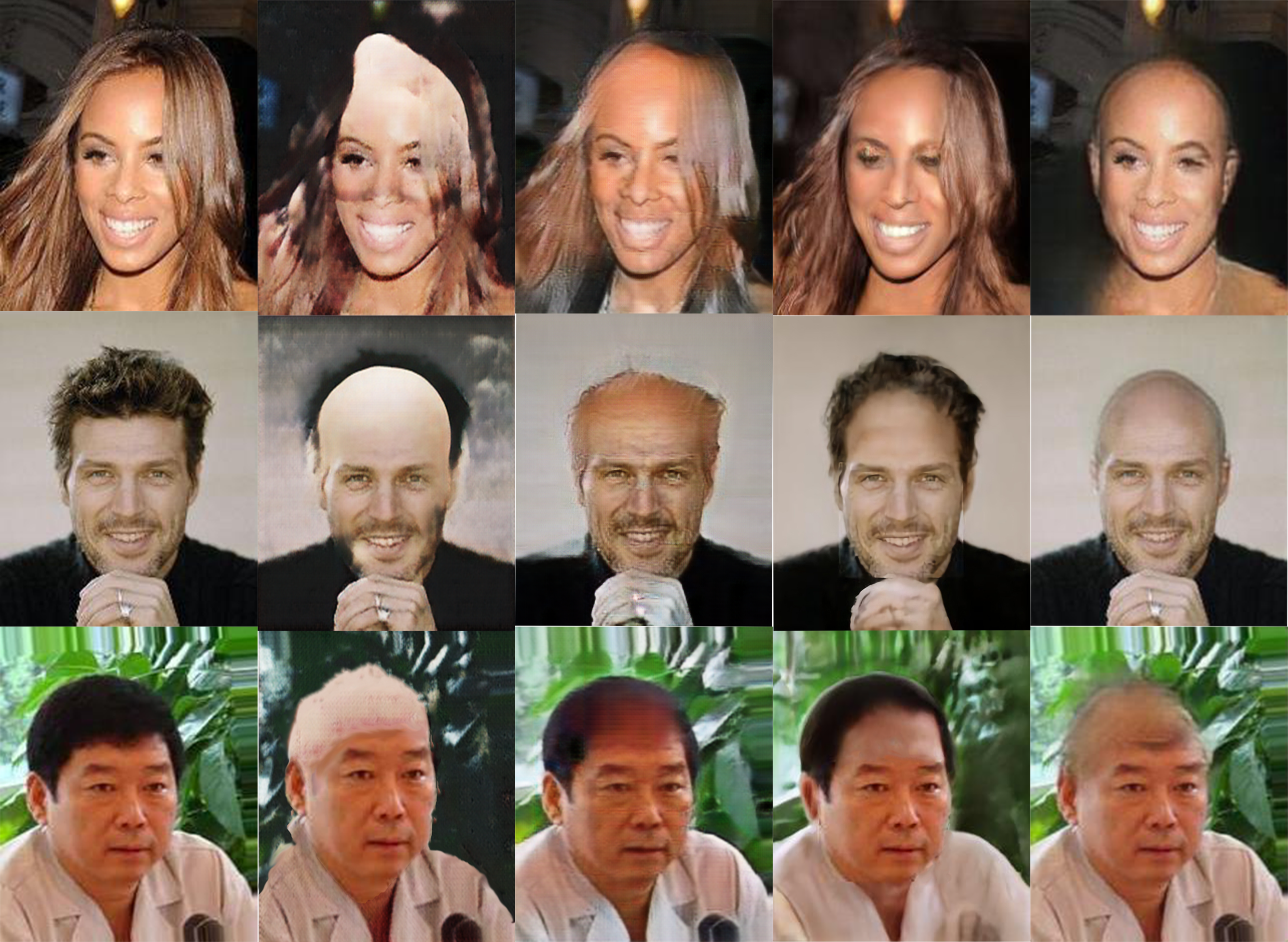}
 \caption{ Comparison of hair removal methods, from left to right: Pix2Pix, CycleGAN, latent space editing and GAN inpainting. The first three methods fail to remove the hair before generating plausible skin, while the inpainting methods offers a realistic reconstruction.
}
\label{fig:QualEvalBald}
\end{figure}

\textbf{Hair removal: }Figure \ref{fig:QualEvalBald} compares the outputs of image translation, latent space editing, and GAN inpainting. The Pix2Pix network which is trained on only synthetic data does not replicate the skin tone of the subject. CycleGAN reconstructs the skin with a more suitable skin tone but superimposes estimated skin on top without removing the hair. Latent space editing works better than the previous methods for men but does not extend well to women. Further, editing the attribute of baldness causes the attribute of age to increase along with increasing the masculinity of women by introducing facial hair. Inpainting gives the best results both visually and generalizing the best to in-the-wild images. It successfully removes the hair and reconstructs skin with an accurate texture that seamlessly blends into the rest of the image.



\begin{figure}[]
\centering
\includegraphics[width = 6.8cm]{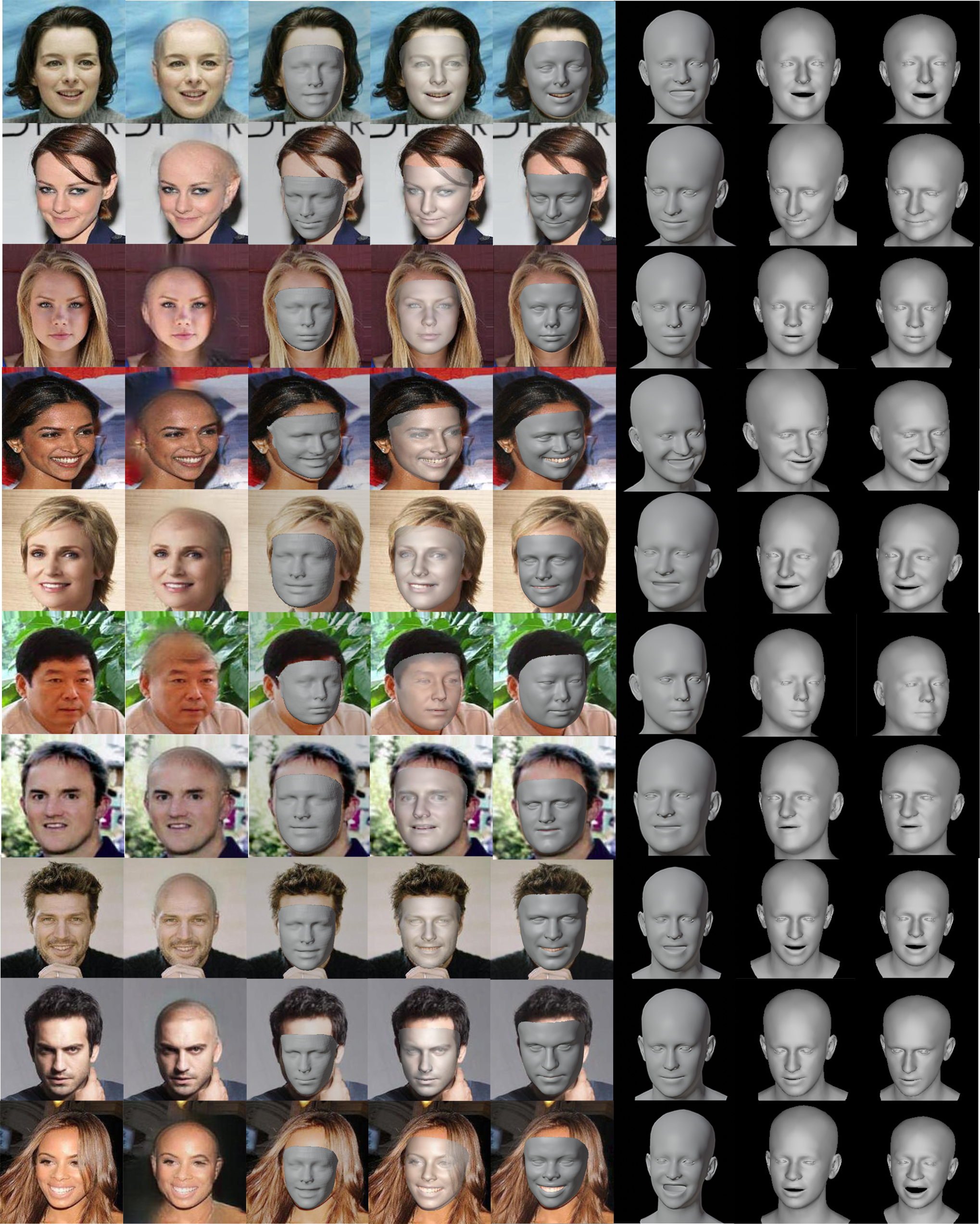}
 \caption{Comparison of reconstruction methods, from left to right: Inpainted image, PRNet\cite{prnet}, 3DDFA-V2\cite{3DDFAV2}, Deep3DRecon\cite{microsoft}, RingNet\cite{ringnet}, DECA\cite{DECA}, and ours. Input images are taken from the CelebA\cite{celebA} dataset.
}
\label{fig:DECA_compare}
\end{figure}

\textbf{Monocular reconstruction: }Figure \ref{fig:DECA_compare} qualitatively compares our method with PRNet\cite{prnet}, 3DDFA-V2\cite{3DDFAV2}, Deep3DRecon\cite{microsoft}, RingNet\cite{ringnet}, and DECA\cite{DECA}. We re-optimize the pose and camera parameters of RingNet and DECA to rigidly align the mesh to match the input image. PRNet, Deep3DRecon, and 3DDFA-V2 do not provide a full head representation while DECA and RingNet geometric shapes have an enlarged head compared to the input image. DECA has better facial details due to the generation of a detail displacement map from the input image. However, this paper focuses on upper head shape reconstruction accuracy and hence does not include this detail map during comparisons.



\begin{figure}[!htb]
\centering
\includegraphics[width = 5.3cm]{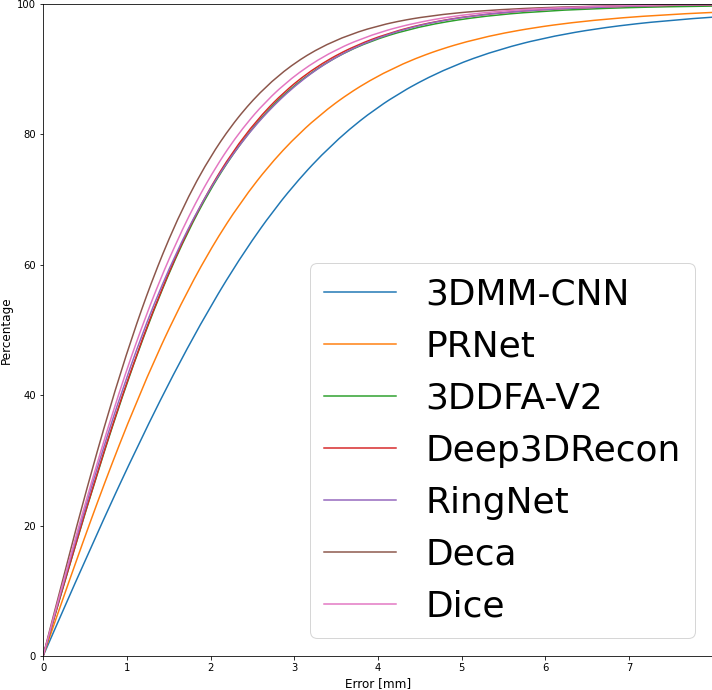}
\caption{Quantitative evaluation on the NoW benchmark\cite{NoWchallenge}}
\label{fig:plot_compare}
\end{figure}

\subsection{Quantitative evaluation}

\subsubsection{Hair removal \label{Hair_removal_result}}




We develop a binary hair vs bald classifier with a ResNet\cite{resnet} architecture trained on the CelebA\cite{celebA} dataset with an accuracy of 98\%. We then compare the percentage of correctly classified images that were artificially manipulated as bald. These results are summarised in table \ref{tab:image_comp_table}. The inpainting network achieves the best accuracy with the classifier and is qualitatively better.

\begin{table}[]
\caption{Comparison of hair removal methods}
\label{tab:image_comp_table}
\centering
\resizebox{\columnwidth}{0.75cm}{%
\begin{tabular}{l c c c }
\hline
\textbf{Method} & \textbf{Number of images} & \textbf{Accuracy (\%) }\\\hline
Inpainting        &20,000 & \textbf{81} \\
CycleGAN        &20,000 & 20 \\
Latent editing        &20,000 & 25 \\
\hline
\end{tabular}
}

\end{table}

\subsubsection{Monocular reconstruction}

We compare the accuracy of facial reconstruction with publicly available methods, namely DECA\cite{DECA} 3DDFAV2\cite{3DDFAV2}, RingNet\cite{ringnet}, PRNet\cite{prnet}, 3DMM-CNN \cite{3DMM_CNN} and Deep3DRecon\cite{microsoft} using the NoW benchmark\cite{NoWchallenge} in Table \ref{tab:all_now_error}, and upper head accuracy on the CoMA\cite{coma} and LYHM\cite{LYHM} dataset in Table \ref{tab:coma_table}.

\textbf{NoW benchmark: }Table \ref{tab:all_now_error} and the cumulative error plot in Figure \ref{fig:plot_compare} compares our results to other popular reconstruction methods on the NoW test set. We have also shown the influence of different cropping levels in table \ref{tab:now_scale_comp}. Although the dice loss offers lower reconstruction error when the entire head is visible, addition of the scale loss results in a more stable reconstruction.


 
\begin{table}[]
\caption{Reconstruction error on the NoW\cite{NoWchallenge} test set}
\label{tab:all_now_error}
\centering

\resizebox{\columnwidth}{!}{%

\begin{tabular}{l c c c }
\hline
\textbf{Method} & \textbf{Median (mm)} & \textbf{Mean (mm )}& \textbf{Std (mm) }\\\hline
3DMM-CNN\cite{3DMM_CNN}        &1.844 &2.33  & 2.04 \\
PRNet\cite{prnet}        &1.499 & 1.976 & 1.88 \\
3DDFA-V2\cite{3DDFAV2}        &1.233 & 1.565 & 1.391 \\
Deep3DRecon\cite{microsoft}        &1.228 & 1.537 & 1.292 \\
RingNet\cite{ringnet}       &1.2065 & 1.534 & 1.306 \\
Dice (\textbf{ours})       &\textbf{1.16} & \textbf{1.47} & \textbf{1.24} \\
DECA\cite{DECA}       &1.09 & 1.38 & 1.18 \\

\hline
\end{tabular}
}

\end{table}




\begin{table}[]
\caption{Comparing reconstruction error on the NoW\cite{NoWchallenge} validation set over all crop levels (fig \ref{fig:cropLevels}). We compare DECA, our model with only dice loss, and our model including scale loss.}
\label{tab:now_scale_comp}
\centering


\begin{tabular}{l c c c }
\hline
\textbf{Method} & \textbf{Median (mm)} & \textbf{Mean (mm )}& \textbf{Std (mm) }\\\hline

\textbf{Crop level 1} & & &\\

DECA\cite{DECA}       &\textbf{1.1785} & \textbf{1.464} & \textbf{1.253} \\
Scale (\textbf{ours})        &{1.206} & {1.507} & {1.274} \\
Dice (ours)       &1.252 & 1.616 & 1.493 \\



\textbf{Crop level 3} & & &\\

DECA\cite{DECA}       &1.293 & 1.641 & 1.424 \\
Scale (\textbf{ours})        &1.161 & 1.445 & 1.209 \\
Dice (\textbf{ours})       &\textbf{1.132} & \textbf{1.409} & \textbf{1.174} \\

\hline
\end{tabular}

\end{table}











\textbf{Full head benchmark: } Our solution aims to solve the problem of accurate reconstruction of the complete head, whereas the NoW benchmark only evaluates metrics on a tightly cropped portion of the face. Thus, we devise an evaluation scheme based on the CoMA\cite{coma} and LYHM\cite{LYHM} dataset. We rigidly align the predicted mesh to the refined ground truth similar to the NoW evaluation and calculate the mean square error on a subset of vertices  corresponding to the upper head between the predicted mesh and the ground truth. Table \ref{tab:coma_table} summarises the error metrics.

\begin{table}[]
\caption{Reconstruction error on the CoMA\cite{coma} and LYHM\cite{LYHM} dataset. We compute the error metric only on vertices corresponding to the top of the head.}
\label{tab:coma_table}
\centering

\resizebox{\columnwidth}{1.25cm}{%

\begin{tabular}{l c c c }
\hline
\textbf{Method} & \textbf{Median (mm)} & \textbf{Mean (mm )}& \textbf{Std (mm) }\\\hline
\textbf{CoMA}\\

DECA        &0.135 &0.136 &0.027  \\
Dice (\textbf{ours})       &\textbf{0.114} &\textbf{0.117} &\textbf{0.025}  \\

\textbf{LYHM}\\

DECA        &3.141 &3.154 &0.142  \\
Dice (\textbf{ours})       &\textbf{3.058} &\textbf{3.048
} &\textbf{0.133}  \\

\hline
\end{tabular}
}

\end{table}

\section{Conclusion}

We address the challenging problem of accurate monocular head reconstruction without 2D-3D supervision. Existing approaches do not explicitly monitor the overall head, causing ambiguity in its shape. To resolve this, we train a stacked hourglass neural network to detect additional landmarks on top of the head for defining its structure. We augment this with a dice consistency loss between the rendered and input images, enabling us to control the fullness of the face and overall shape when detected landmarks do not perfectly align with facial contours. Since the rendered image does not have hair, we require the subject in the input image to be bald for accurate computation of the overlap degree. After experimenting with image translation, latent space editing, and inpainting methods, we utilise a hypergraph-based inpainting network for hair removal and skin reconstruction. This network also learns to reconstruct ears which were previously occluded by hair in the input images. Future work can use the ears in the reconstruction pipeline to improve the 3D pose and shape accuracy. However, this method can produce minor artifacts in the image, especially when long hair obscures the face or in profile images, suggesting the need to train on a more challenging dataset. Further, we introduce a training method to account for different crops of the input image. However, the scale loss improves stability at the expense of accuracy, suggesting the need for more varied training data. Our model is built upon FLAME\cite{FLAME} and hence, is compatible with DECA's\cite{DECA} displacement map generation pipeline. Model weights corresponding to DECA's detail network are directly transferable to our model. We present a complete framework that regresses accurate head geometry and facial features from monocular in-the-wild images for use in 3D avatar creation.




\acknowledgments{
The first author would like to thank the India Connect program for the opportunity to conduct research at Nanyang Technological University. We would also like to thank S. Sanyal and Yao Feng for their RingNet and DECA pytorch implementation.
}
\bibliographystyle{abbrv-doi}

\bibliography{template}

\begin{thebibliography}{10}

\bibitem{faceparsing}
Face parsing using the bisenet architecture.
\newblock \url{https://github.com/zllrunning/face-parsing.PyTorch}, 2018.

\bibitem{NoWchallenge}
{\em NoW challenge}, 2019.

\bibitem{blanzvetter}
V.~Blanz and T.~Vetter.
\newblock A morphable model for the synthesis of 3d faces.
\newblock {\em SIGGRAPH'99 Proceedings of the 26th annual conference on
  Computer graphics and interactive techniques}, 09 2002. doi: {{%
10\hspace{.1pt}\discretionary{.}{%
}{.}\hspace{.4pt}1145\discretionary{/}{%
}{/}311535\hspace{.1pt}\discretionary{.}{%
}{.}\hspace{.4pt}311556}}


\bibitem{FAN}
A.~Bulat and G.~Tzimiropoulos.
\newblock How far are we from solving the 2d {\&} 3d face alignment problem?
  (and a dataset of 230, 000 3d facial landmarks).
\newblock {\em CoRR}, abs/1703.07332, 2017.

\bibitem{vggface2}
Q.~Cao, L.~Shen, W.~Xie, O.~M. Parkhi, and A.~Zisserman.
\newblock Vggface2: {A} dataset for recognising faces across pose and age.
\newblock {\em CoRR}, abs/1710.08092, 2017.

\bibitem{blender}
B.~O. Community.
\newblock {\em Blender - a 3D modelling and rendering package}.
\newblock Blender Foundation, 2018.

\bibitem{LYHM}
H.~Dai, N.~Pears, W.~Smith, and C.~Duncan.
\newblock Statistical modeling of craniofacial shape and texture.
\newblock {\em International Journal of Computer Vision}, 128(2):547--571, Nov
  2019. doi: {{%
10\hspace{.1pt}\discretionary{.}{%
}{.}\hspace{.4pt}1007\discretionary{/}{%
}{/}s11263\discretionary{%
}{-}{-}019\discretionary{%
}{-}{-}01260\discretionary{%
}{-}{-}7}}


\bibitem{microsoft}
Y.~Deng, J.~Yang, S.~Xu, D.~Chen, Y.~Jia, and X.~Tong.
\newblock Accurate 3d face reconstruction with weakly-supervised learning: From
  single image to image set, 2020.

\bibitem{DECA}
Y.~Feng, H.~Feng, M.~J. Black, and T.~Bolkart.
\newblock Learning an animatable detailed 3d face model from in-the-wild
  images, 2021.

\bibitem{prnet}
Y.~Feng, F.~Wu, X.~Shao, Y.~Wang, and X.~Zhou.
\newblock Joint 3d face reconstruction and dense alignment with position map
  regression network.
\newblock In {\em ECCV}, 2018.

\bibitem{ganfit}
B.~Gecer, S.~Ploumpis, I.~Kotsia, and S.~Zafeiriou.
\newblock Ganfit: Generative adversarial network fitting for high fidelity 3d
  face reconstruction.
\newblock {\em 2019 IEEE/CVF Conference on Computer Vision and Pattern
  Recognition (CVPR)}, Jun 2019. doi: {{%
10\hspace{.1pt}\discretionary{.}{%
}{.}\hspace{.4pt}1109\discretionary{/}{%
}{/}cvpr\hspace{.1pt}\discretionary{.}{%
}{.}\hspace{.4pt}2019\hspace{.1pt}\discretionary{.}{%
}{.}\hspace{.4pt}00125}}


\bibitem{genova}
K.~Genova, F.~Cole, A.~Maschinot, A.~Sarna, D.~Vlasic, and W.~T. Freeman.
\newblock Unsupervised training for 3d morphable model regression.
\newblock {\em CoRR}, abs/1806.06098, 2018.

\bibitem{GAN}
I.~Goodfellow, J.~Pouget-Abadie, M.~Mirza, B.~Xu, D.~Warde-Farley, S.~Ozair,
  A.~Courville, and Y.~Bengio.
\newblock Generative adversarial nets.
\newblock {\em Advances in neural information processing systems}, 27, 2014.

\bibitem{3DDFAV2}
J.~Guo, X.~Zhu, Y.~Yang, F.~Yang, Z.~Lei, and S.~Z. Li.
\newblock Towards fast, accurate and stable 3d dense face alignment.
\newblock {\em CoRR}, abs/2009.09960, 2020.

\bibitem{resnet}
K.~He, X.~Zhang, S.~Ren, and J.~Sun.
\newblock Identity mappings in deep residual networks.
\newblock {\em CoRR}, abs/1603.05027, 2016.

\bibitem{psnr}
A.~Horé and D.~Ziou.
\newblock Image quality metrics: Psnr vs. ssim.
\newblock In {\em 2010 20th International Conference on Pattern Recognition},
  pp. 2366--2369, 2010. doi: {{%
10\hspace{.1pt}\discretionary{.}{%
}{.}\hspace{.4pt}1109\discretionary{/}{%
}{/}ICPR\hspace{.1pt}\discretionary{.}{%
}{.}\hspace{.4pt}2010\hspace{.1pt}\discretionary{.}{%
}{.}\hspace{.4pt}579}}


\bibitem{dynamicavatar}
A.~Ichim, S.~Bouaziz, and M.~Pauly.
\newblock Dynamic 3d avatar creation from hand-held video input.
\newblock {\em ACM Transactions on Graphics}, 34:45:1--45:14, 07 2015. doi: {{%
10\hspace{.1pt}\discretionary{.}{%
}{.}\hspace{.4pt}1145\discretionary{/}{%
}{/}2766974}}


\bibitem{globallocal}
S.~Iizuka, E.~Simo-Serra, and H.~Ishikawa.
\newblock Globally and locally consistent image completion.
\newblock {\em ACM Transactions on Graphics}, 36:1--14, 07 2017. doi: {{%
10\hspace{.1pt}\discretionary{.}{%
}{.}\hspace{.4pt}1145\discretionary{/}{%
}{/}3072959\hspace{.1pt}\discretionary{.}{%
}{.}\hspace{.4pt}3073659}}


\bibitem{pix2pix}
P.~Isola, J.-Y. Zhu, T.~Zhou, and A.~A. Efros.
\newblock Image-to-image translation with conditional adversarial networks,
  2018.

\bibitem{stylegan}
T.~Karras, S.~Laine, and T.~Aila.
\newblock A style-based generator architecture for generative adversarial
  networks, 2019.

\bibitem{celebAMaskHQ}
C.-H. Lee, Z.~Liu, L.~Wu, and P.~Luo.
\newblock Maskgan: Towards diverse and interactive facial image manipulation.
\newblock In {\em IEEE Conference on Computer Vision and Pattern Recognition
  (CVPR)}, 2020.

\bibitem{FLAME}
T.~Li, T.~Bolkart, M.~J. Black, H.~Li, and J.~Romero.
\newblock Learning a model of facial shape and expression from 4d scans.
\newblock {\em ACM Transactions on Graphics (TOG)}, 36:1 -- 17, 2017.

\bibitem{partial_conv}
G.~Liu, F.~A. Reda, K.~J. Shih, T.~Wang, A.~Tao, and B.~Catanzaro.
\newblock Image inpainting for irregular holes using partial convolutions.
\newblock {\em CoRR}, abs/1804.07723, 2018.

\bibitem{celebA}
Z.~Liu, P.~Luo, X.~Wang, and X.~Tang.
\newblock Deep learning face attributes in the wild.
\newblock In {\em Proceedings of International Conference on Computer Vision
  (ICCV)}, December 2015.

\bibitem{edge_connect}
K.~Nazeri, E.~Ng, T.~Joseph, F.~Z. Qureshi, and M.~Ebrahimi.
\newblock Edgeconnect: Generative image inpainting with adversarial edge
  learning.
\newblock {\em CoRR}, abs/1901.00212, 2019.

\bibitem{stackedhourglass}
A.~Newell, K.~Yang, and J.~Deng.
\newblock Stacked hourglass networks for human pose estimation.
\newblock {\em CoRR}, abs/1603.06937, 2016.

\bibitem{coma}
A.~Ranjan, T.~Bolkart, S.~Sanyal, and M.~J. Black.
\newblock Generating {3D} faces using convolutional mesh autoencoders.
\newblock In {\em European Conference on Computer Vision (ECCV)}, pp. 725--741,
  2018.

\bibitem{ringnet}
S.~Sanyal, T.~Bolkart, H.~Feng, and M.~J. Black.
\newblock Learning to regress 3d face shape and expression from an image
  without 3d supervision, 2019.

\bibitem{interfacegan}
Y.~Shen, C.~Yang, X.~Tang, and B.~Zhou.
\newblock Interfacegan: Interpreting the disentangled face representation
  learned by gans.
\newblock {\em TPAMI}, 2020.

\bibitem{totalmoving}
S.~Suwajanakorn, I.~Kemelmacher-Shlizerman, and S.~M. Seitz.
\newblock Total moving face reconstruction.
\newblock In {\em ECCV}, 2014.

\bibitem{face2face}
J.~Thies, M.~Zollhöfer, M.~Stamminger, C.~Theobalt, and M.~Nießner.
\newblock Face2face: Real-time face capture and reenactment of rgb videos,
  2020.

\bibitem{3DMM_CNN}
A.~T. Tran, T.~Hassner, I.~Masi, and G.~G. Medioni.
\newblock Regressing robust and discriminative 3d morphable models with a very
  deep neural network.
\newblock {\em CoRR}, abs/1612.04904, 2016.

\bibitem{Hypergraphs}
G.~Wadhwa, A.~Dhall, S.~Murala, and U.~Tariq.
\newblock Hyperrealistic image inpainting with hypergraphs, 2020.

\bibitem{msssim}
Z.~Wang, E.~Simoncelli, and A.~Bovik.
\newblock Multiscale structural similarity for image quality assessment.
\newblock In {\em The Thrity-Seventh Asilomar Conference on Signals, Systems
  Computers, 2003}, vol.~2, pp. 1398--1402 Vol.2, 2003. doi: {{%
10\hspace{.1pt}\discretionary{.}{%
}{.}\hspace{.4pt}1109\discretionary{/}{%
}{/}ACSSC\hspace{.1pt}\discretionary{.}{%
}{.}\hspace{.4pt}2003\hspace{.1pt}\discretionary{.}{%
}{.}\hspace{.4pt}1292216}}


\bibitem{inversion_survey}
W.~Xia, Y.~Zhang, Y.~Yang, J.~Xue, B.~Zhou, and M.~Yang.
\newblock {GAN} inversion: {A} survey.
\newblock {\em CoRR}, abs/2101.05278, 2021.

\bibitem{bisenet}
C.~Yu, J.~Wang, C.~Peng, C.~Gao, G.~Yu, and N.~Sang.
\newblock Bisenet: Bilateral segmentation network for real-time semantic
  segmentation, 2018.

\bibitem{GNN_segment}
K.~Zhang, T.~Li, S.~Shen, B.~Liu, J.~Chen, and Q.~Liu.
\newblock Adaptive graph convolutional network with attention graph clustering
  for co-saliency detection.
\newblock {\em CoRR}, abs/2003.06167, 2020.

\bibitem{lpips}
R.~Zhang, P.~Isola, A.~A. Efros, E.~Shechtman, and O.~Wang.
\newblock The unreasonable effectiveness of deep features as a perceptual
  metric.
\newblock {\em CoRR}, abs/1801.03924, 2018.

\bibitem{indomain}
J.~Zhu, Y.~Shen, D.~Zhao, and B.~Zhou.
\newblock In-domain gan inversion for real image editing.
\newblock In {\em Proceedings of European Conference on Computer Vision
  (ECCV)}, 2020.

\bibitem{cyclegan}
J.-Y. Zhu, T.~Park, P.~Isola, and A.~A. Efros.
\newblock Unpaired image-to-image translation using cycle-consistent
  adversarial networks, 2020.

\bibitem{stateoftheart}
M.~Zollhöfer, J.~Thies, P.~Garrido, D.~Bradley, T.~Beeler, P.~Pérez,
  M.~Stamminger, M.~Nießner, and C.~Theobalt.
\newblock State of the art on monocular 3d face reconstruction, tracking, and
  applications.
\newblock {\em Computer Graphics Forum}, 37:523--550, 05 2018. doi: {{%
10\hspace{.1pt}\discretionary{.}{%
}{.}\hspace{.4pt}1111\discretionary{/}{%
}{/}cgf\hspace{.1pt}\discretionary{.}{%
}{.}\hspace{.4pt}13382}}


\end{thebibliography}
\clearpage


\section{Appendix}

\begin{figure*}[!htb]

\centering

\begin{subfigure}{.18\textwidth}
\includegraphics[ width = \textwidth]{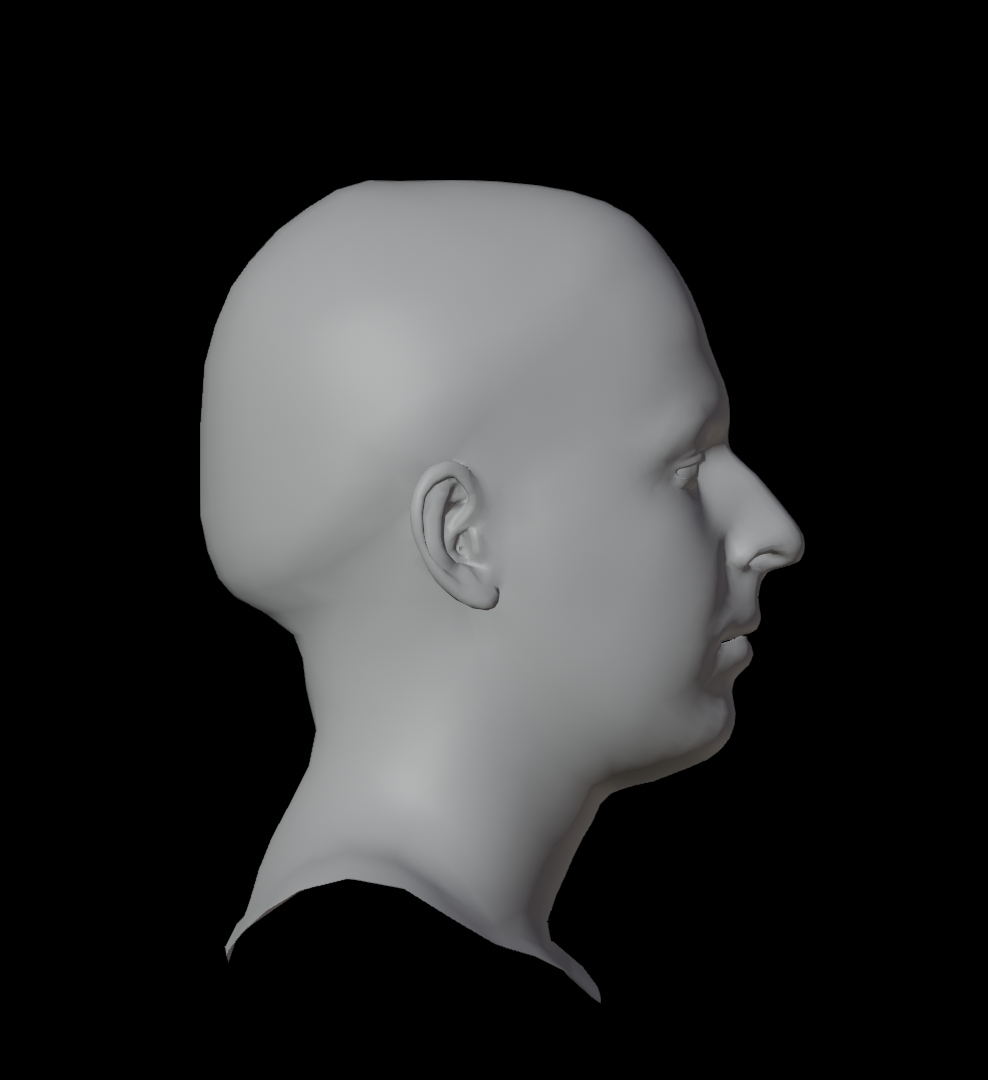}
\caption{Original CoMA FLAME registration with hair cap baked into the mesh. }
\label{c_a}
\end{subfigure}
~
\begin{subfigure}{.18\textwidth}
\includegraphics[ width=\textwidth]{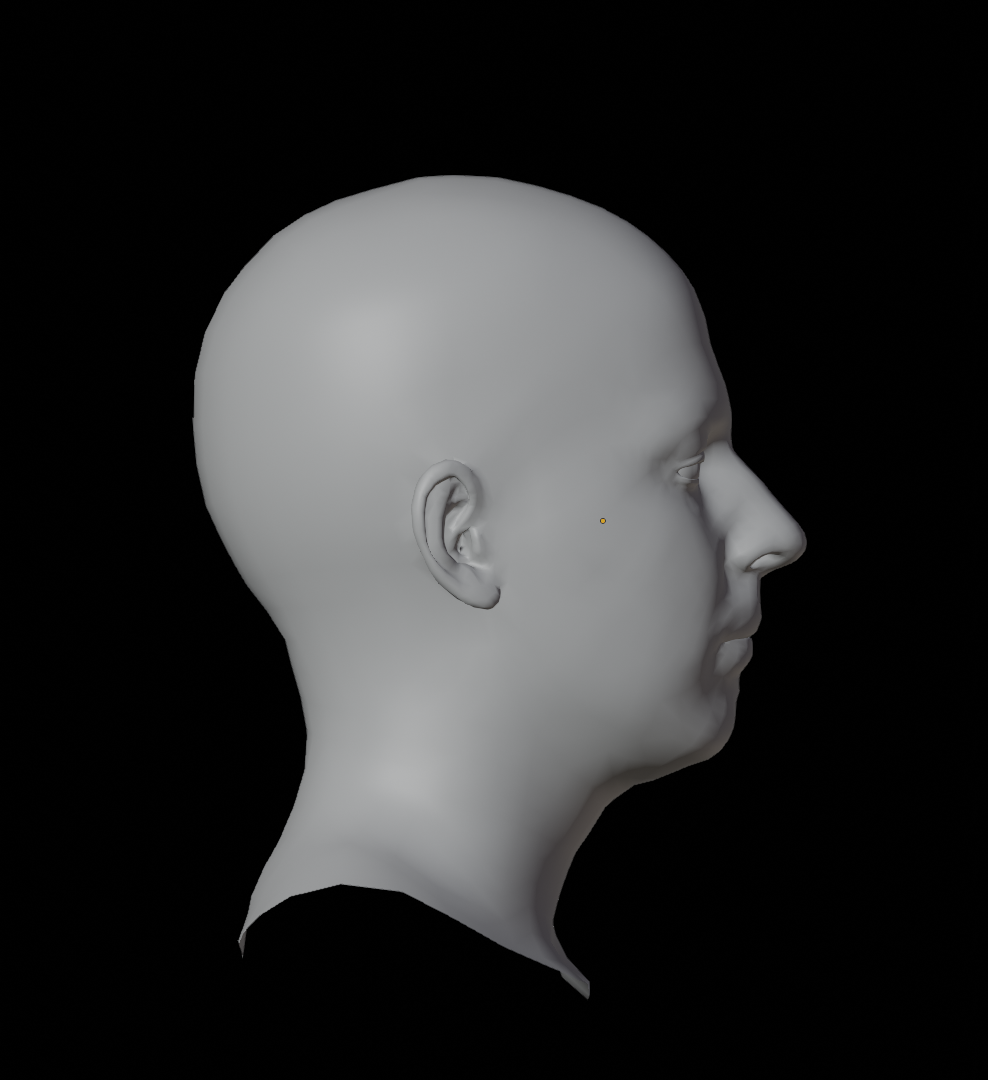}
\caption{Mesh in the FLAME topology after refitting with a lower tolerance.}
\label{c_b}
\end{subfigure}
~
\begin{subfigure}{.18\textwidth}
\includegraphics[ width=\textwidth]{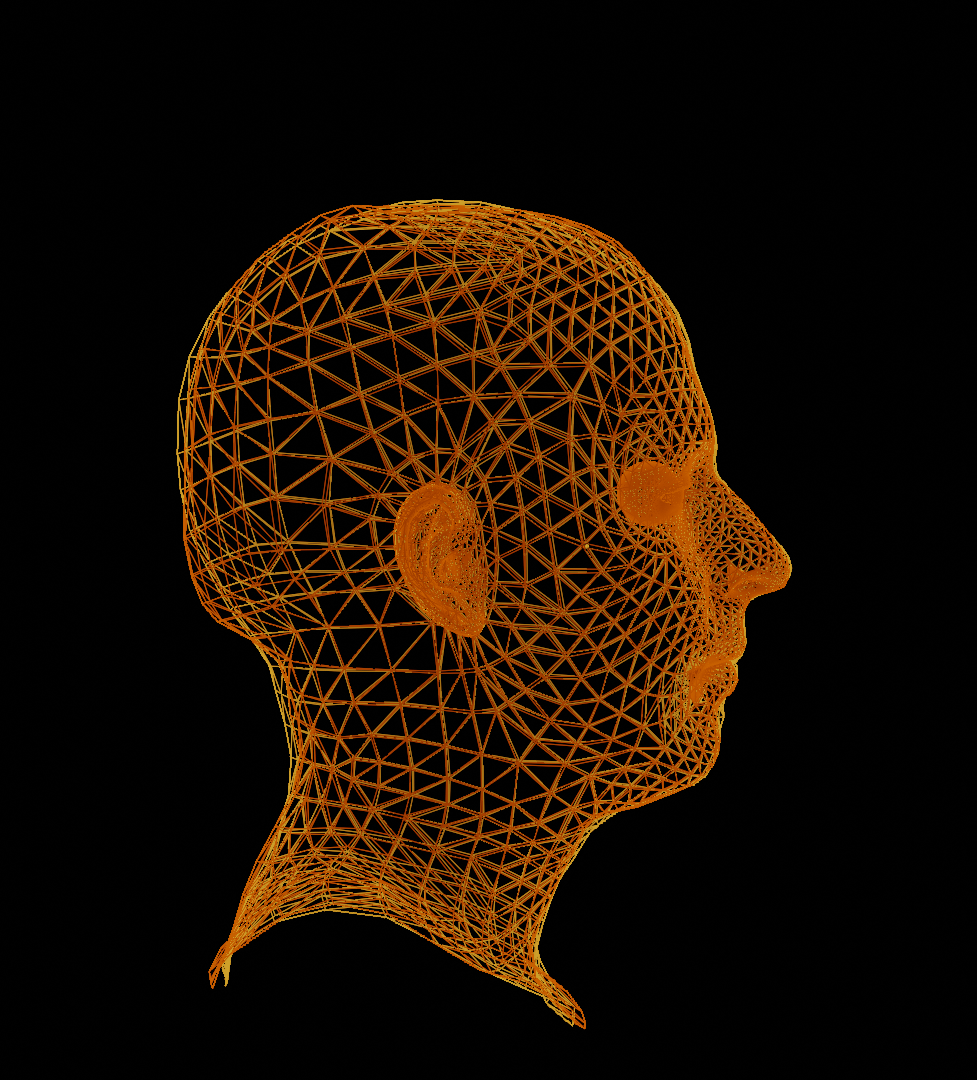}
\caption{Comparing the refit FLAME geometry (yellow) to the original mesh (orange).}
\label{c_c}
\end{subfigure}
~
\begin{subfigure}{.18\textwidth}
\includegraphics[ width=\textwidth]{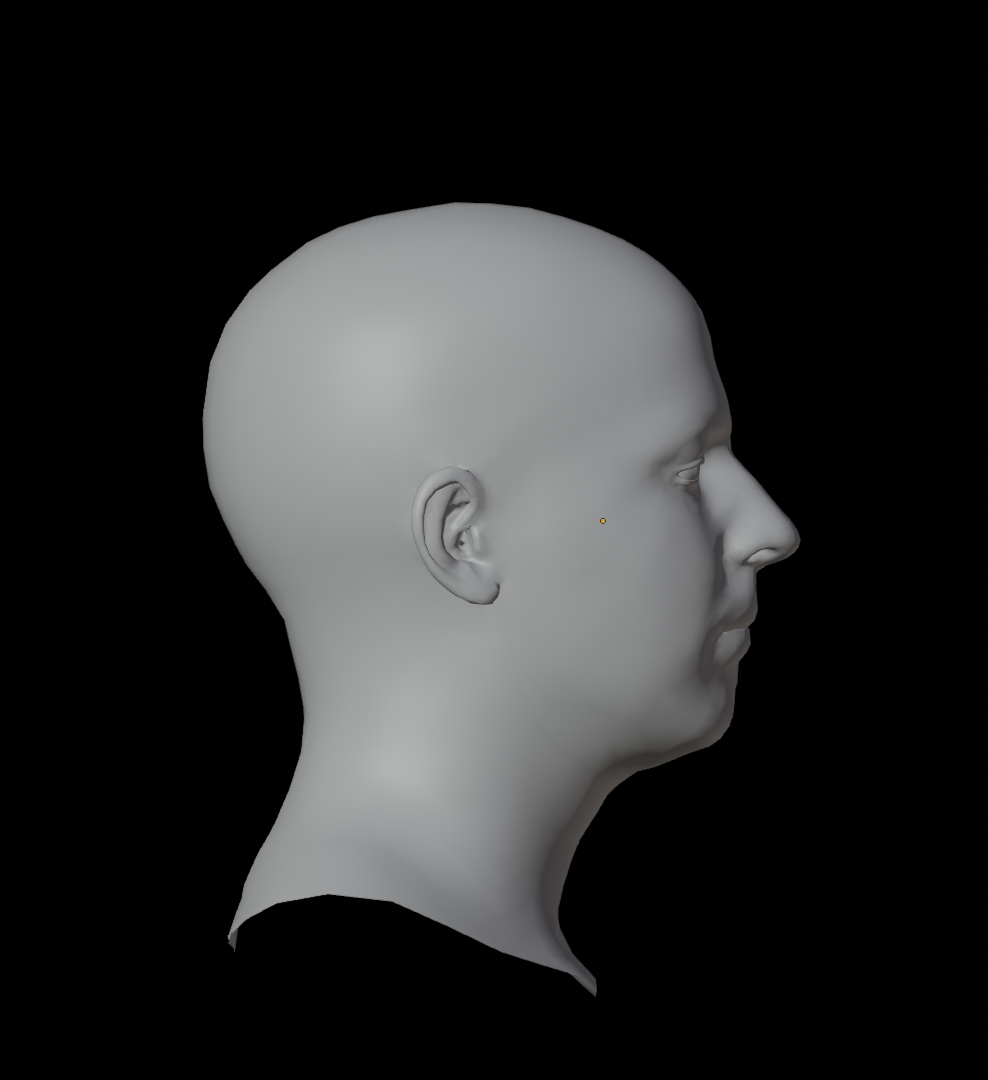}
\caption{Manually deforming the refitted mesh to obtain better representation of shape.}
\label{c_d}
\end{subfigure}
~
\begin{subfigure}{.18\textwidth}
\includegraphics[ width=\textwidth]{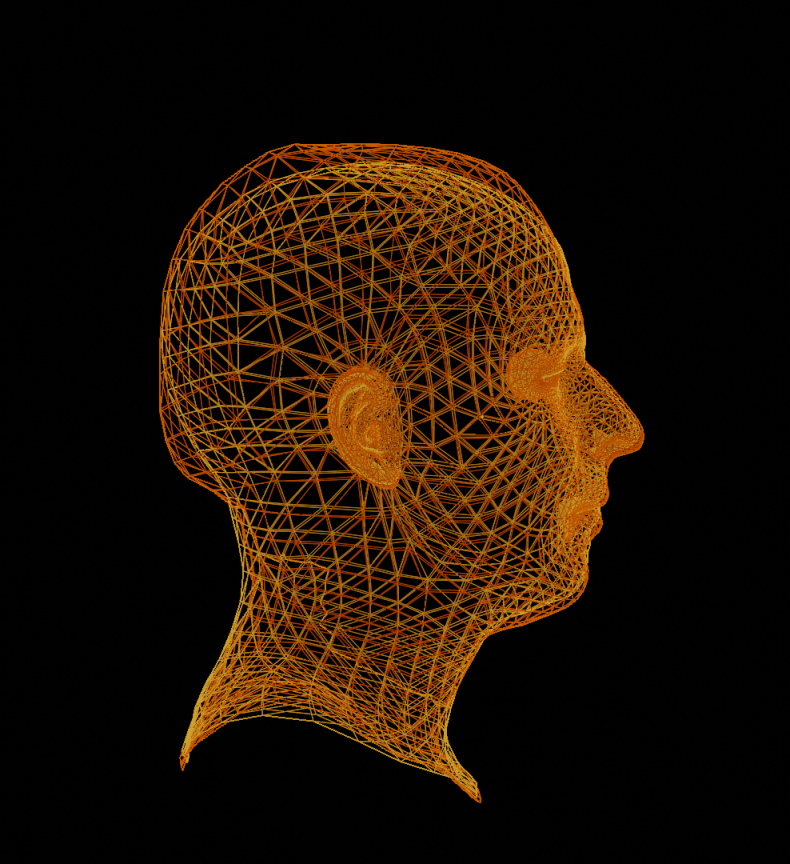}
\caption{Comparing manually refined mesh (yellow) to original mesh (orange).}
\label{c_e}
\end{subfigure}
~
\caption{Steps involved in cleanup of the CoMA dataset.}

\label{all_plots}
\end{figure*}

\textbf{A. CoMA Benchmark}\\
\\
The CoMA dataset consists of 3D scans of 12 extreme, asymmetric facial expressions from 12 different subjects. We use this dataset to evaluate upper head accuracy. However, the provided FLAME registrations also include geometric deformations corresponding to the hair cap used during scanning (Fig. \ref{c_a}), leading to inaccurate ground truth head shape. We clean the meshes using the following method:

1) For each subject We fit FLAME again to the provided FLAME registrations, but with a looser tolerance as shown in Figure \ref{c_b}. 

2) The new FLAME fitting offers a smoother head shape and removes the deformations due to hair mask, but the resultant head is of similar size to the original mesh with hair cap (Fig. \ref{c_c}). Thus, we manually refine the neutral head shape of a subject in Blender (Fig. \ref{c_d}) and compute the non-rigid transformation between the manually deformed mesh and FLAME refitting.  

3) This non-rigid transformation is applied to all the new fittings of the subject resulting in refined meshes for all expressions of a particular subject. These meshes are used to compute the error metric.\\

~

\noindent \textbf{B. More qualitative results}\\

\begin{figure*}[]
\centering

\includegraphics[width = \linewidth]{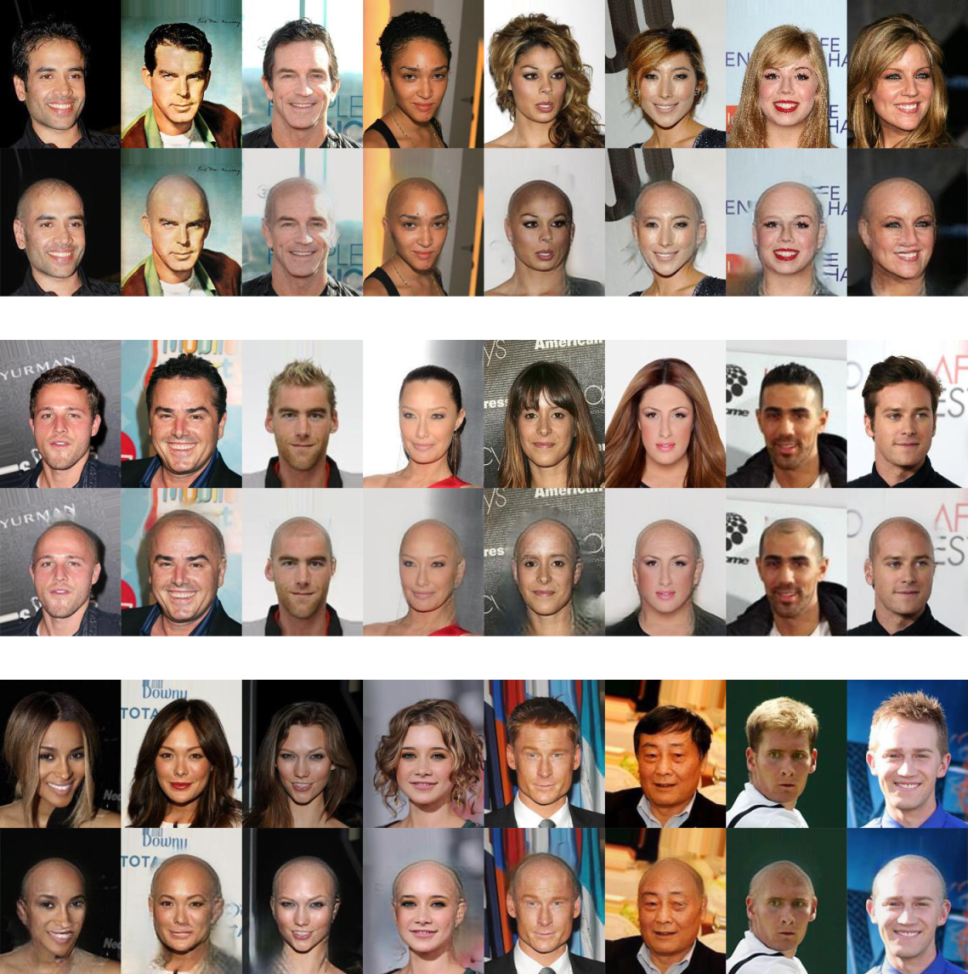}
 \caption{Inpainting results, top: input images taken from CelebA dataset, bottom: inpainted output. Inpainting is robust to image conditions like luminance and contrast. It also reconstructs the ear that was occluded by hair in the input.
}
\label{fig:inpaint1}
\end{figure*}

\begin{figure*}[]
\centering

\includegraphics[width = \linewidth]{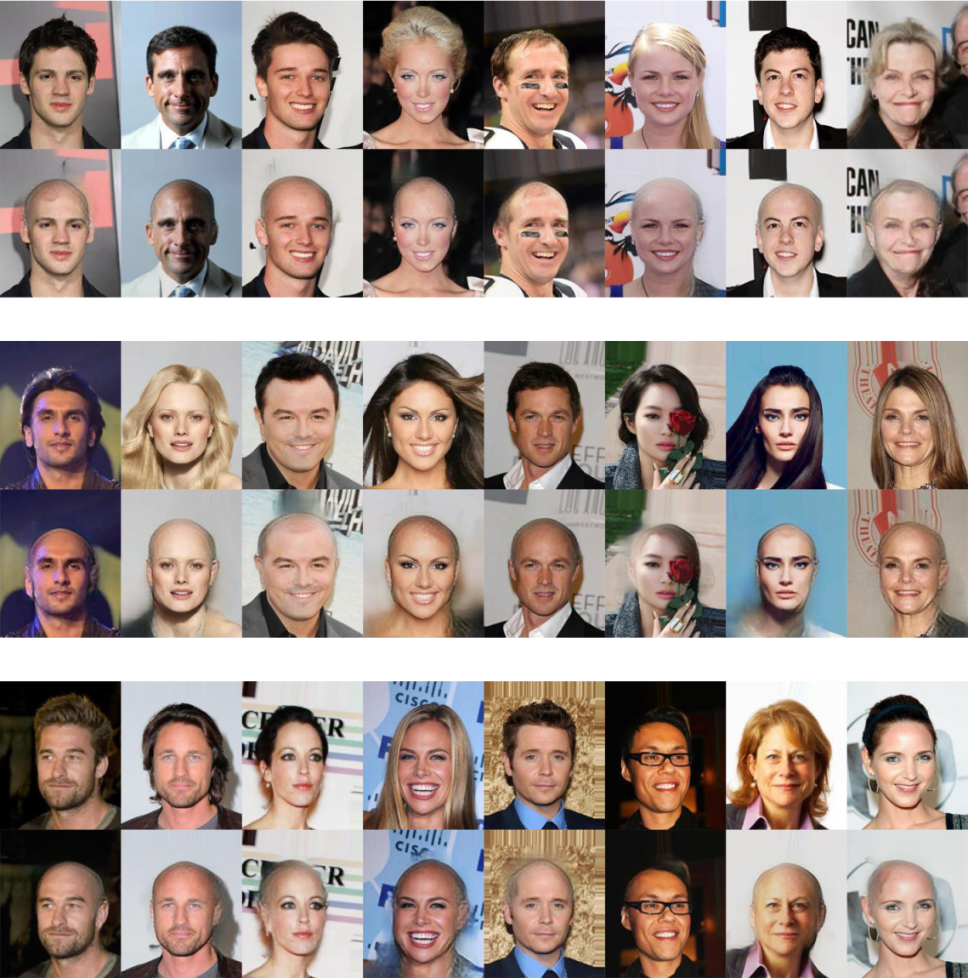}
 \caption{ Inpainting results, top: input images taken from CelebA dataset, bottom: inpainted output. Inpainting is robust to image conditions like luminance and contrast. It also reconstructs the ear that was occluded by hair in the input.
}
\label{fig:inpaint2}
\end{figure*}

\begin{figure*}[]
\centering

\includegraphics[width = 0.8\linewidth ]{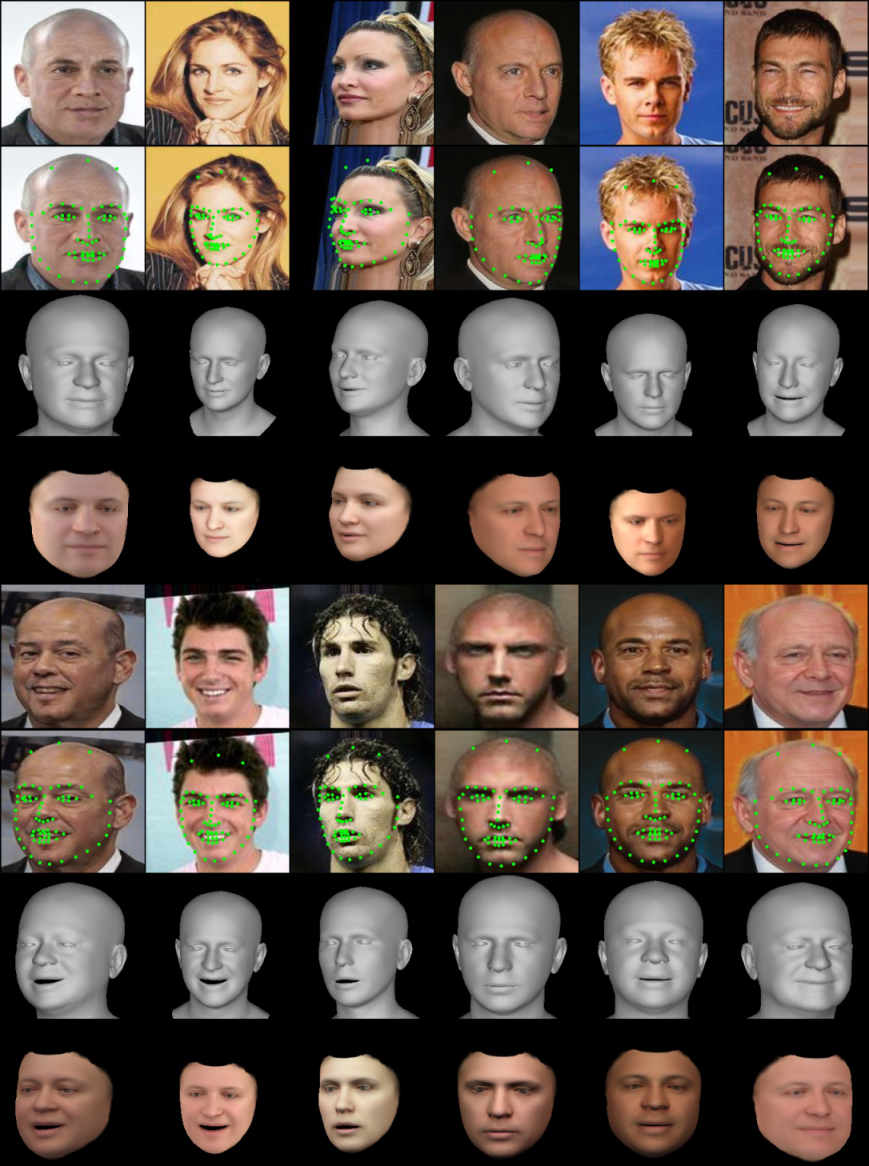}
 \caption{From top to bottom: input images from CelebA dataset, ground truth landmarks, projected landmarks, reconstructed shape, texture.
}
\label{fig:reco1}
\end{figure*}

\begin{figure*}[]
\centering

\includegraphics[width = 0.8\linewidth]{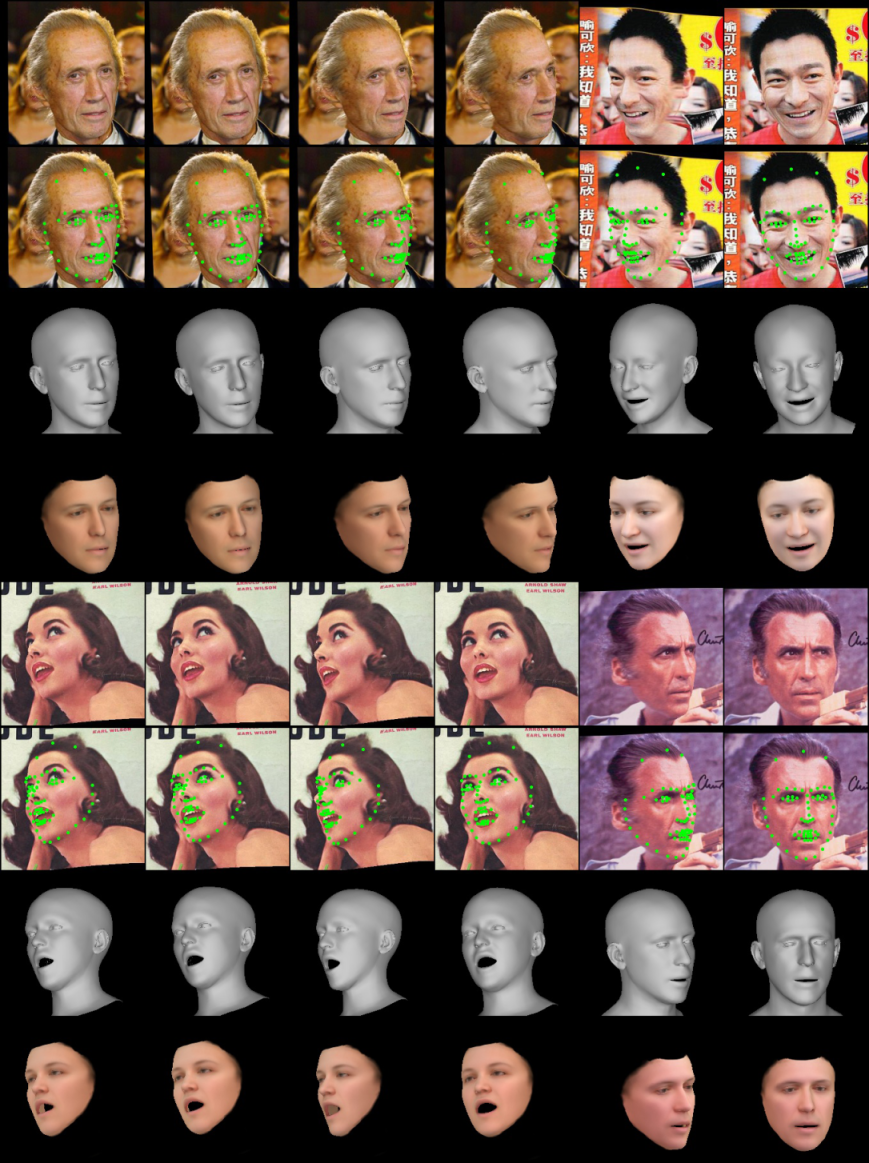}
 \caption{Illustrating the scale loss where images are in varying pose, from top to bottom: input images from LFPW dataset, ground truth landmarks, projected landmarks, reconstructed shape, texture.
}
\label{fig:reco3}
\end{figure*}

\begin{figure*}[]
\centering

\includegraphics[width = 0.8\linewidth]{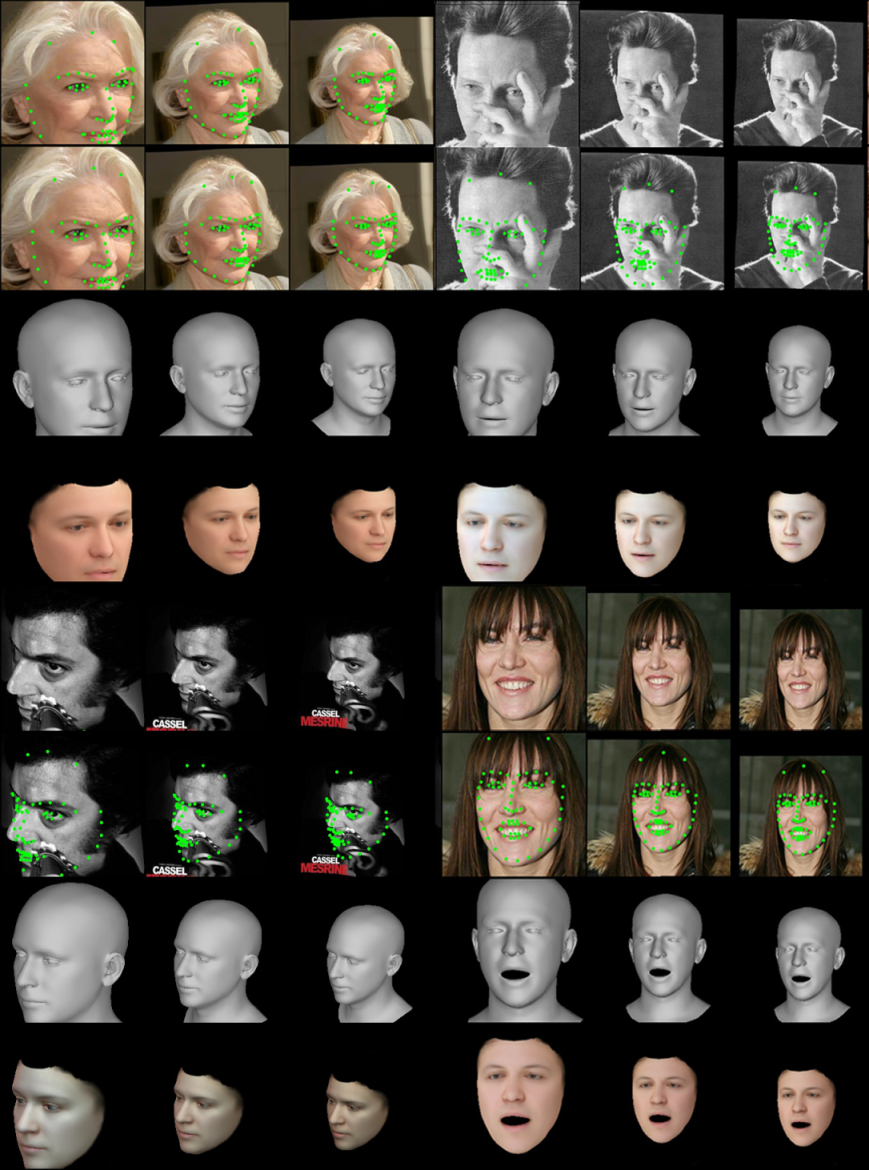}
 \caption{Illustrating the scale loss with 3 scale levels, from top to bottom: input images from LFPW dataset, ground truth landmarks, projected landmarks, reconstructed shape, texture.
}
\label{fig:reco4}
\end{figure*}

\begin{figure*}[]
\centering

\includegraphics[height = 0.8\textheight]{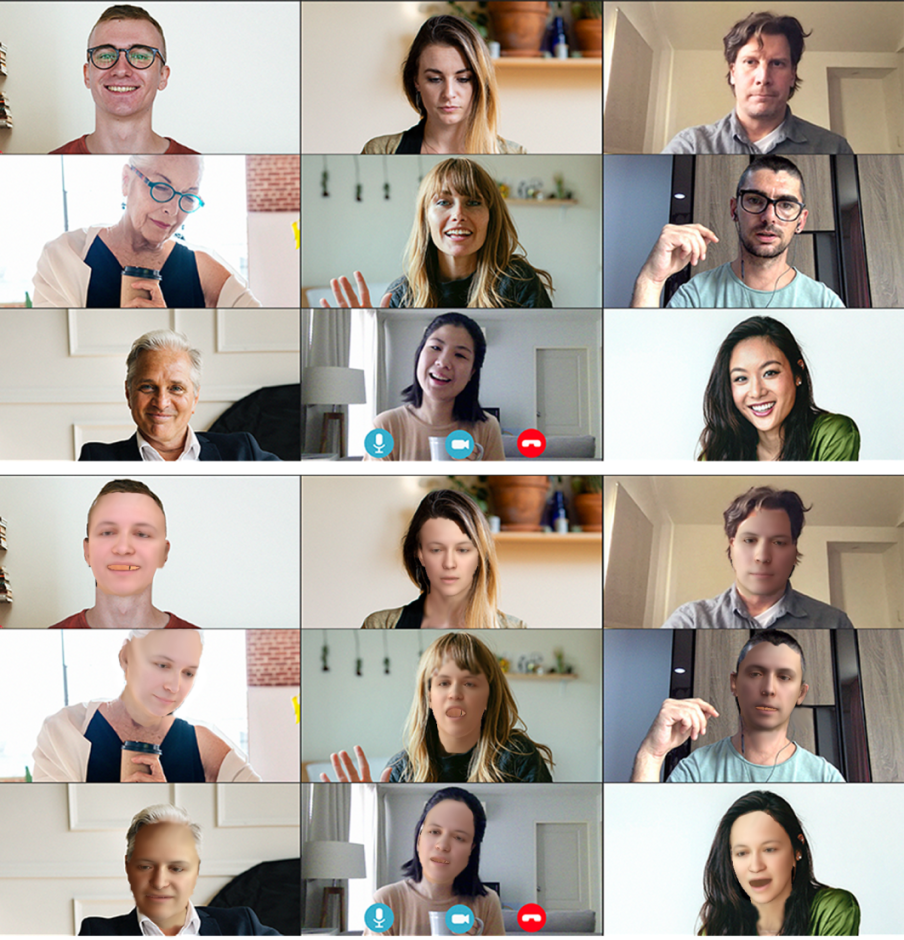}
 \caption{ A potential application of our reconstruction method. Top: example input of a video call. Bottom: reconstructed shape and rendered using FLAME texture model. 
}
\label{fig:meeting}
\end{figure*}

 \noindent Figures \ref{fig:inpaint1} and \ref{fig:inpaint2} show additional inpainting results on the CelebA dataset. The hypergraph-based inpainting approach removes hair effectively regardless of gender and hair length. Figures \ref{fig:reco1}, \ref{fig:reco3}, and \ref{fig:reco4} show additional qualitative results of monocular 3D shape reconstruction. Figures \ref{fig:reco1} demonstrate the first half of training where only dice loss is utilized, while Figures \ref{fig:reco3} and \ref{fig:reco4} show the latter half when we introduce the scale loss. These input images have varying poses and scales. The reconstructions closely follow the input image's facial contours and have an accurate head shape even when the upper part of the head is not visible. A potential application of our method is in remote collaboration. Compared to current video-based conferencing, personalized avatars can be used for meetings in a virtual environment. Figure \ref{fig:meeting} shows the results of our method on a screenshot of a conference call.


\end{document}